\pdfoutput=1
\documentclass[11pt]{article}

\usepackage[final]{acl}

\usepackage{times}
\usepackage{latexsym}
\usepackage{todonotes}
\usepackage{inconsolata}
\usepackage{multirow, makecell}
\usepackage{scrextend}
\usepackage{array}
\usepackage{arydshln}
\usepackage{float}
\usepackage{tikz}
\usepackage{floatpag}
\usepackage{csquotes}
\usepackage{ltablex}
\usepackage{caption}

\usepackage{tabularray}
\UseTblrLibrary{booktabs}

\usepackage[T1]{fontenc}
\usepackage[utf8]{inputenc}

\usepackage{microtype}

\usepackage{inconsolata}

\newcommand\blfootnote[1]{%
  \begingroup
  \renewcommand\thefootnote{}\footnote{#1}%
  \addtocounter{footnote}{-1}%
  \endgroup
}

\usepackage{array}
\newcolumntype{L}[1]{>{\raggedright\let\newline\\\arraybackslash\hspace{0pt}}m{#1}}
\newcolumntype{C}[1]{>{\centering\let\newline\\\arraybackslash\hspace{0pt}}m{#1}}
\newcolumntype{R}[1]{>{\raggedleft\let\newline\\\arraybackslash\hspace{0pt}}m{#1}}

\title{Are Emergent Abilities in Large Language Models just In-Context Learning?}

\author{
Sheng Lu\textsuperscript{1*},
Irina Bigoulaeva\textsuperscript{1*},
Rachneet Sachdeva\textsuperscript{1}, \\
\textbf{Harish Tayyar Madabushi}\textsuperscript{2},
\and
\textbf{Iryna Gurevych\textsuperscript{1}} \\ [0.3cm]
\textsuperscript{1} Ubiquitous Knowledge Processing Lab, Technical University of Darmstadt \\
\textsuperscript{2} Department of Computer Science, The University of Bath \\
\texttt{\small www.ukp.tu-darmstadt.de} \\ [-0.1mm]
\texttt{\small htm43@bath.ac.uk} \\
}

\begin{document}
\maketitle
\blfootnote{\hspace*{-0.112cm}\textsuperscript{*}Equal Contribution.

Accepted to ACL 2024. A longer version of this paper is available at \url{https://h-tayyarmadabushi.github.io/Emergent_Abilities_and_in-Context_Learning/}.}
\begin{abstract}
Large language models, comprising billions of parameters and pre-trained on extensive web-scale corpora, have been claimed to acquire certain capabilities without having been specifically trained on them. These capabilities, referred to as ``emergent abilities,'' have been a driving force in discussions regarding the potentials and risks of language models. A key challenge in evaluating emergent abilities is that they are confounded by model competencies that arise through alternative prompting techniques, including in-context learning, which is the ability of models to complete a task based on a few examples. We present a novel theory that explains emergent abilities, taking into account their potential confounding factors, and rigorously substantiate this theory through over 1000 experiments. Our findings suggest that purported emergent abilities are not truly emergent, but result from a combination of in-context learning, model memory, and linguistic knowledge. Our work is a foundational step in explaining language model performance, providing a template for their efficient use and clarifying the paradox of their ability to excel in some instances while faltering in others. Thus, we demonstrate that their capabilities should not be overestimated.~\footnote{Our code and data are available at \url{https://github.com/UKPLab/on-emergence} and \url{https://tudatalib.ulb.tu-darmstadt.de/handle/tudatalib/3931}.}
\end{abstract}

% \setcounter{tocdepth}{3}
% \tableofcontents
% \pagebreak
\section{Introduction, Motivation and Context}

One of the most captivating aspects of pre-trained language models (PLMs) is their capacity to acquire a wide range of knowledge across different domains, while being trained primarily through masked language modelling, a task requiring models to predict masked tokens in their input 
~\cite{tenney2018what,petroni-etal-2019-language}.
The diverse abilities of PLMs can be categorised into two broad types: formal linguistic abilities and functional linguistic abilities. Formal linguistic abilities refer to the understanding of language rules and patterns, which PLMs, for example, BERT~\cite{devlin-etal-2019-bert} are known to excel at~\cite{tenney2018what,petroni-etal-2019-language}. The latter category includes a range of abilities akin to human cognition that are necessary for real-world language use and comprehension, such as commonsense knowledge and social awareness. While PLMs excel at formal linguistic abilities, they have faced challenges in developing functional linguistic abilities~\cite{DBLP:journals/corr/abs-2301-06627}. 

The introduction of Large Language Models (LLMs), which are typically generative PLMs scaled up to billions of parameters and trained on vast, web-scale data corpora, is changing this landscape \cite{DBLP:conf/nips/BrownMRSKDNSSAA20,DBLP:journals/jmlr/ChowdheryNDBMRBCSGSSTMRBTSPRDHPBAI23,DBLP:journals/corr/abs-2302-13971,DBLP:journals/corr/abs-2307-09288}. Recent works indicate that LLMs exhibit \emph{emergent abilities}, as measured by their above random performance without explicit training on tasks, including those tasks that explicitly require some form of reasoning. An emergent ability was first defined as an ability to solve a task which is absent in smaller models, but present in LLMs. This definition, introduced approximately concurrently by two works~\cite{wei2022emergent,srivastava2023beyond}, is based on the more general definition of emergence in physics: ``Emergence is when quantitative changes in a system result in qualitative changes in behaviour''~\cite{anderson1972more}. Emergent abilities are implied due to LLMs' capacity to perform above the random baseline on the corresponding tasks without explicit training on those same tasks. For example, the emergent ability to understand social situations in LLMs is inferred from LLMs' performing well above the random baseline on the Social IQA~\cite{sap-etal-2019-social} task, which serves to evaluate models' emotional and social intelligence and includes questions such as ``Carson was excited to wake up to attend school. Why did he do this? Options: Take the big test, Go to bed early, Just say hello to friend (correct)''. 

\subsection{Significance for Applications and Safety}
\label{sec:intro-safty}
While prior work on emergent abilities does not explicitly make the distinction between formal and functional linguistic abilities, the identification of numerous functional linguistic capabilities holds profound implications for both the potential and safety of LLMs. The assumption that LLMs have access to emergent functional linguistic abilities significantly affects the way in which users interact with and use these systems. Overreliance on these perceived abilities can lead users to provide insufficiently detailed instructions, potentially resulting in hallucinations and errors. If there are indeed multiple functional linguistic abilities that emerge with scale, it suggests that further scaling has the potential to unlock a wide array of additional abilities which we cannot predict, especially since they tend not to present themselves in smaller-scale models~\cite{wei2022emergent}. This inherent unpredictability associated with emergent abilities holds substantial implications for the discussion surrounding safety and security when utilising LLMs. Indeed, it has been argued that these could include potentially hazardous abilities, 
including reasoning and planning~\cite{DBLP:journals/ais/Hoffmann23}, thereby posing an existential threat to humanity~\cite{bengio2023managing}. In this work, we refer to such potentially harmful capabilities, as ``latent hazardous abilities.''

It's important to emphasise that the development of linguistic proficiencies (i.e. formal linguistic abilities) does \emph{not} carry the potentials of this nature. The same can be said for the capacity to efficiently handle information retrieval tasks. The real focus lies on potential capabilities relating to functional linguistic abilities. However, it must be emphasised that this does not include other dangers posed through the misuse of these models, such as the use of LLMs to generate fake news. Similarly, we do not contend that future AI systems could \emph{never} pose an existential threat. Instead, we clarify that, contrary to prevailing narratives, the evidence from LLM abilities does \emph{not} support this concern.

\subsection{Abilities vs. Techniques}
\label{sec:subsection-promptingtech}
The scaling up of LLMs facilitates the acquisition of diverse competencies, which can be grouped into two categories: The first encompasses \emph{abilities}, already described. The second encompasses various \emph{techniques}, which LLMs can benefit from. These techniques show less of an effect in smaller models, but become progressively more effective with scale. Among these techniques are in-context learning and instruction-tuning. In-context learning (ICL) is the technique wherein LLMs are provided with a limited number of examples within the input prompt itself~\cite{DBLP:conf/nips/BrownMRSKDNSSAA20}. From these examples, the model infers how to perform a specific task, responding appropriately to the question posed by the prompt~\cite{DBLP:conf/nips/BrownMRSKDNSSAA20,DBLP:journals/csur/LiuYFJHN23}.
Investigations into the theoretical underpinnings of ICL and its specific manifestation in LLMs indicate that it might bear resemblance to the process of fine-tuning models on the specific tasks for which they are provided examples~\cite{DBLP:conf/iclr/AkyurekSA0Z23,dai2023why,DBLP:conf/icml/OswaldNRSMZV23,wei2023larger}. Another technique exclusive to LLMs is instructional fine-tuning, alternatively known as instruction-tuning. This technique involves fine-tuning LLMs on datasets of prompts and their corresponding desired outputs, which enables the models to follow explicit instructions in prompts~\cite{DBLP:journals/corr/abs-2210-11416,wei2022finetuned,taori2023alpaca}.
%Finally, chain-of-thought prompting is the technique in which models are provided with a sequence of intermediate reasoning steps to boost their `reasoning skills' ~\cite{wei2022chain}. 
Following previous work~\cite{wei2022emergent}, we refer to these techniques, illustrated in Figure \ref{fig:instructTuned}, as \textit{prompting techniques}. 

Significant to our investigation is the observation that prompting techniques and emergent abilities manifest within LLMs at a comparable scale. Furthermore, ICL and instruction-tuning can be observed in smaller-scale models, albeit to a lesser degree, and are thus predictable. This predictability means they are not `emergent', nor do they pose a threat, contrasting with the unpredictability and potential risks associated with emergent abilities in larger models. %As such, it is crucial to emphasise that ``prompting techniques'' do not give rise to safety concerns with respect to latent hazardous abilities because prompting techniques provide a way for users to steer the output of LLMs. 
%Additionally, pinpointing what allows models to solve tasks -- by differentiating user-controllable techniques from the models' inherent functional linguistic capabilities -- promises to improve how individuals effectively use these models.
%predictable outcome that is reliant on the user. 
%These prompting techniques offer novel avenues for leveraging models, but cannot on their own lead to the development of latent hazardous abilities.
%Hence, this manner of risk is exclusively a consequence of emergent abilities. As previously mentioned, this threat is not universal among all emergent abilities, but pertains exclusively to those involving reasoning or planning. 
%To be explicit, formal linguistic abilities do not pose any threat, and nor does the ability of models to perform above the random baseline on tasks that can be solved through memory (\textit{`memorisable tasks'}) since such proficiency only indicates models' ability to memorise information. 
Considering this context, 
%Given that prompting techniques manifest themselves at the same scale as the emergence of a significant number of abilities, coupled with the safety and usage implications associated with emergent abilities but not with the prompting techniques, 
it becomes imperative to ascertain the extent of these emergent abilities in the absence of prompting techniques. 

\subsection{Fine-tuning, In-Context Learning, and other Prompting Techniques}
\label{section:ft-in-context-prompts}
Artificial neural models have, for some time, exhibited tremendous success on specific tasks when trained on those tasks \cite{devlin-etal-2019-bert,DBLP:journals/corr/abs-1907-11692}. PLMs in particular have demonstrated this even when trained on just a few examples~\cite{DBLP:journals/corr/abs-1811-05468,radford2019language,DBLP:conf/nips/BrownMRSKDNSSAA20,DBLP:conf/acl/GaoFC20}. Such performance is not considered ``emergent'', precisely because models are trained on that very task. Indeed, the fact that LLMs are \emph{not} trained on the tasks used in evaluating their emergent abilities is central to identifying abilities which are \emph{truly} emergent. 
The assertion that achieving satisfactory performance on a given task signifies the \emph{emergence} of associated `abilities' hinges on the condition that models are not explicitly trained for that specific task.

The recent insights indicating parallels between ICL and explicit training suggest that the success on a task through ICL, much like models trained explicitly for task-solving, does not imply a model inherently possessing that \emph{ability}~\cite{dai2023why}. For example, it has been shown that ICL implements gradient descent implicitly and constructs a function at inference time on regression problems \cite{DBLP:conf/iclr/AkyurekSA0Z23,li2023transformers,DBLP:journals/corr/abs-2306-09927}, which may be related to gradient-based meta-learning \cite{DBLP:conf/icml/OswaldNRSMZV23}. Importantly, however, the specific mechanisms governing ICL do not impact our argument: The fact of its functionality suffices to underscore the necessity of assessing emergent abilities in the absence of ICL.
%Instructional tuning is another prompting technique that has garnered success in LLMs. 
Additionally, instruction-tuning datasets typically include several variations of an instruction followed by the task input or context (see Figure \ref{fig:instructTuned}). As such, we contend that the process of instruction fine-tuning potentially enables models to map prompts to in-context examples (detailed in Section \ref{sec:t5performance}), thereby utilising ICL to respond to prompts. This would imply that the success of a model to solve a task in this scenario also does not indicate the emergence of the corresponding ability.

The safety issues associated with LLMs stem from their ability to perform well above the random baseline on tasks that cannot be solved through memorisation and are indicative of certain `abilities', \emph{without explicit training on those tasks}. Therefore, recognising that prompts act as a form of `training mechanism' rather than simply a way of interfacing with a model with inherent functional linguistic abilities offers the potential to alter how we use these models and deepen our understanding of their capabilities and limitations. As such, it is crucial to conduct an independent evaluation of LLMs' abilities, detached from ICL.

\subsection{Research Questions and Contributions} 
\label{sec:researchquestions}
Our research seeks to answer two pivotal questions: Firstly, in light of ICL's influence on perceived emergent abilities in LLMs, which abilities are truly emergent in the absence of ICL, including instructional tuning? Secondly, given LLMs' capability for ICL and the typical inclusion of instruction-exemplar mappings in instruction-tuning datasets, can we find evidence of the emergence of functional linguistic abilities in instruction-tuned models? Or can ICL better explain their capabilities and shortcomings?

Our primary contribution lies in demonstrating the absence of emergent functional linguistic abilities in LLMs when ICL is not a factor, thus demystifying the true capabilities of LLMs and affirming their safety, while additionally dispelling concerns over potential latent hazardous abilities. Our secondary contributions include empirically testing the hypothesis that instruction-tuned models' capabilities stem from efficient ICL, thus offering an explanation for LLMs' abilities as stemming from a combination of formal linguistic skills, vast information retention and recall, and notably, ICL. By identifying user-directable ICL, rather than intrinsic functional linguistic capabilities, as the mechanism behind LLM performance, we lay out a framework for more efficient use of these models, shedding light on their capabilities and limitations.% including why they excel in certain cases but fail in others.

\section{Experimental Setup}
\label{sec:experiments-main}
In this section, we present an overview of our experimental methods investigating emergent abilities in the absence of ICL. We experiment with 20 models across 22 tasks using two different settings. We use four different evaluation metrics and additionally run multiple tests for bias, including a manual analysis of our results. We present an overview of this setup below, while details on the hyperparameters and training regime are presented in Appendix \ref{app:experimental-setup}.

\subsection{Models}
\label{sec:models}
We experiment with four model families: GPT, T5~\cite{2020t5}, Falcon\footnote{See \url{https://falconllm.tii.ae/index.html}.} and LLaMA~\cite{DBLP:journals/corr/abs-2302-13971}. We choose these model families, since GPT and LLaMA have previously been found to have emergent abilities, and Falcon is at the top of LLM leaderboards at the time of writing. Finally, we select T5 as it is an encoder-decoder model, and its instruction-tuned version (Flan) is trained using an extensive instruction-tuning dataset. Table \ref{tab:model_details} enumerates the models that we use in our experiments. The emergence of abilities in relation to scale requires the evaluation of each model family across a range of sizes (parameter counts), and so we select models at different scales from each of these families. Important to our inquiry is the hypothesis that instructional tuning might indirectly leverage ICL. In light of this possibility, we experiment with both.

\begin{table}[h!]
\small
\centering
\begin{tabular}{llc}
\toprule
\textbf{Model}                    & \textbf{Instruction-Tuned Version} & \textbf{Size} \\ \midrule
GPT-2                             & GPT-2-IT                           & 117M       \\
GPT-2-XL                          & GPT-2-XL-IT                        & 1.6B       \\
GPT-J                             & GPT-JT                             & 6.7B       \\
\multirow{2}{*}{\texttt{davinci}} & \texttt{text-davinci-001}          & \multirow{2}{*}{175B} \\
                                  & \texttt{text-davinci-003}          &            \\ \midrule
T5-small                          & Flan-T5-small                      & 60M        \\
T5-large                          & Flan-T5-large                      & 770M       \\ \midrule
Falcon-7B                         & Falcon-7B-Instruct                 & 7B         \\
Falcon-40B                        & Falcon-40B-Instruct                & 40B        \\ \midrule
LLaMA-7B                          & --                                 & 7B         \\
LLaMA-13B                         & --                                 & 13B        \\
LLaMA-30B                         & --                                 & 30B        \\
\bottomrule
\end{tabular}
\caption{\label{tab:model_details} Details of the models used in the experiments.} % InstrucTuned refers to the instruction fine-tuned version of the model. \emph{text-davinci-003} is andditionally trained using RLHF.}
\end{table}

\subsection{Tasks}
\label{sec:tasks}
In selecting tasks to assess the emergence of abilities, we base our selection on those tasks that have been identified as emergent in GPT-3 by prior works. We refer to these tasks as \emph{previously identified as emergent}. Out of 17 such tasks in the BIG-bench dataset~\cite{srivastava2023beyond}, we incorporate 14 into our study. Three tasks previously identified as emergent are excluded from our analysis, because their generative nature made them challenging to assess automatically in a manner consistent with the other tasks. Additionally, to create a baseline for comparison, we randomly choose seven tasks from the same dataset that were not previously identified as emergent. Finally, we also include GSM8K \cite{DBLP:journals/corr/abs-2110-14168}, which comprises a set of grade-school mathematics word problems and is noteworthy because even the latest models struggle with this task. 

Given that formal linguistic abilities and the capacity to efficiently handle information retrieval tasks do not pose an existential threat, we manually analyse the proficiency required to solve each of the tasks we select. A full list of tasks, including their memorisability and classification as functional or formal linguistic abilities, is presented in Table \ref{tab:default-results-ntz}. We determine memorisability through a manual analysis of 50 examples from each task. We provide details of our manual analysis and examples from each task in the Appendix \ref{app:task-memorisability}.

\subsection{Settings}
We evaluate each model on each task using both the few-shot and the zero-shot settings. When using the few-shot setting, we use 5 in-context examples. We note that the few-shot setting explicitly makes use of ICL, whereas the zero-shot setting does not. 

\subsection{Evaluation Metrics}
\label{sec:evalmetrics}
To account for the possibility that the outputs generated by non-instruction-tuned models do not match the provided answer choices exactly, we additionally evaluate using the metric BERTScore accuracy, which calculates the semantic similarity between the output text and the provided answer choices using BERTScore~\cite{DBLP:conf/iclr/ZhangKWWA20} 
%\footnote{BERTScore V 0.3.13 using RoBERTa Large}%, 355M parameters, available at \url{https://huggingface.co/FacebookAI/roberta-large/commit/716877d372b884cad6d419d828bac6c85b3b18d9}} 
to estimate the model's answer choice. In this setting, the answer is considered correct if the generated answer is most similar (semantic text similarity) to the correct answer choice, and incorrect if it is closer to any of the others. The majority of the results we present in our analysis are based on this evaluation metric. It's worth noting that this is akin to selecting the answer where the model has exhibited lowest perplexity. Since calculating this perplexity for models that are exclusively accessible through APIs is not practical, we adopt this alternative metric. We opt for BERTScore over alternatives like BLEURT~\cite{DBLP:conf/acl/SellamDP20} because the latter are additionally trained to assess the fluency of the output text, a factor which is not our focus, and one that renders them computationally resource-intensive. For tasks that require the output of a number or a coded string (i.e., Modified arithmetic, GSM8K, and Codenames), we limit our evaluation to exact matching, as measuring semantic similarity between numbers or coded strings does not accurately reflect their proximity. 

 Additionally, given that recent work has indicated that emergence might be a result of discrete evaluation metrics~\cite{schaeffer2023emergent}, we also include string edit distance. Our investigation reveals that the the lack of emergence is consistent across the metrics we use, and thus we do not use continuous metrics in our analysis. Overall, we evaluate using exact match accuracy, BERTScore accuracy, and string edit distance.

\subsection{Control for Bias and Manual Evaluation}
In order to ensure that our evaluation is fair, we identify potential biases that could influence our findings and design our experiments to mitigate such biases. First, to ensure that non-instruction-tuned models are not disadvantaged by the typically instructional task prompts, we modify these prompts, by refining them to ensure their solvability even in the absence of instruction comprehension. We then experiment with minor variations to these prompts to find the most optimal format. We also experiment with using the shortened output format, where models are only required to output a letter associated with the correct answer. We do this to remove the dependence on the non-exact-match evaluation metrics. Importantly, we manually evaluate the output of our models to ensure that the prompts where appropriately interpreted by the models, especially those which are not instruction tuned.
Details of these experiments and associated results are presented in Appendix \ref{sec:prompts}. 
 
\section{Emergence in GPT in the Absence of In-Context Learning}
\label{sec:results-emerge}

In this and the next section, we highlight a subset of the results with the goal of highlighting the key findings and trends from our experiments. Specifically, this section deals with the emergence of functional linguistic abilities in non-instruction-tuned models, and the next section (Section \ref{sec:t5performance}) focuses on exploring instruction-tuned models and their interplay with ICL and emergent abilities. Considering that prior research has identified emergent abilities in GPT%, and our task selection is influenced by these findings, 
we prioritise the GPT family in our experimental analysis. 

\begin{figure*}[ht!]
\centering
\includegraphics[width=0.9\textwidth]{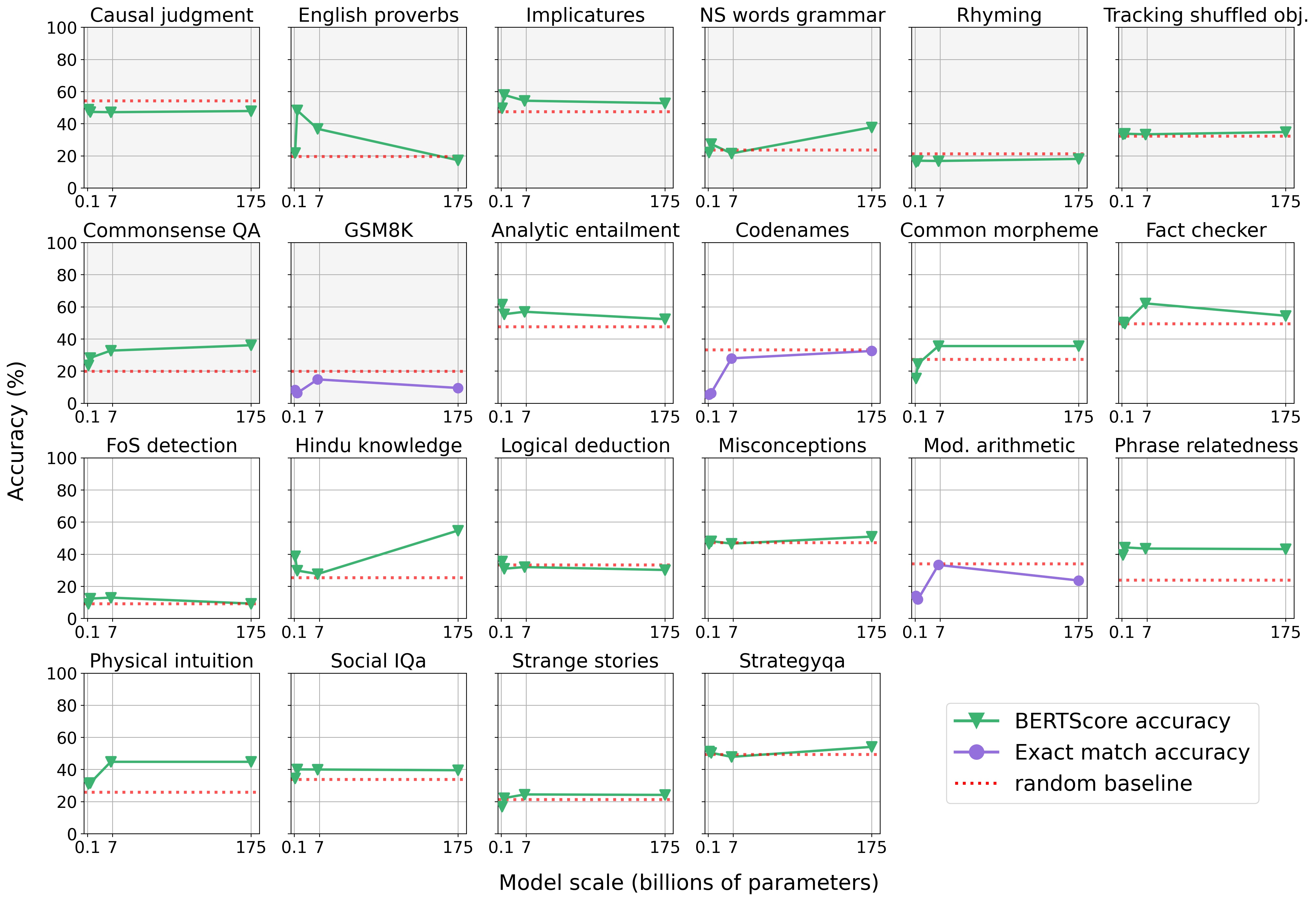}
\caption{\label{fig:gpt-results} Performance of non-instruction-tuned GPT models in the zero-shot setting. Grey background indicates tasks that are not previously identified as emergent. Tasks that require the output of a number or a coded string are evaluated using exact match accuracy. Note the consistent lack of ``emergence'', see text for details.}
\end{figure*}

Figure \ref{fig:gpt-results} illustrates the performance of non-instruction-tuned models from the GPT family in the setting where they are prompted without the use of in-context examples (zero-shot). This approach guarantees the exclusion of ICL, allowing for a clear assessment of emergent abilities. Tasks listed in the first row against a grey background are tasks which have not been found to be emergent by prior work and the rest are those which have been found to be emergent previously.%~\cite{wei2022emergent}. 

\label{sec:no-incontext-performance}

Recall that the definition of emergence~\cite{wei2022emergent} requires LLMs to perform a task above the baseline \emph{and} do so in a manner that cannot be predicted based on the performance of smaller models. An analysis of Figure \ref{fig:gpt-results}, presented in Table \ref{tab:default-results-ntz} indicates that just two tasks are ``emergent'' when we control for ICL. While two additional tasks (Misconceptions and Strategy QA) also have unpredictable above-baseline performance, the improvement is only marginal, as these tasks are binary classification tasks with a random baseline of 50\% accuracy. Among the two identified tasks, Nonsense words grammar pertains to a formal linguistic ability, which we have noted does not involve any latent hazardous abilities such as reasoning. The other task, Hindu knowledge, solely relies on information recall and likewise does not demand any reasoning. As such, we find no functional linguistic abilities emergent in \texttt{davinci}, the non-instruction-tuned 175B GPT model in the absence of ICL.

\begin{table*}[!ht]
\footnotesize
\centering
\begin{tabular}{lccccc}
\toprule
\textbf{Task} & \textbf{Competence Type} & \textbf{Memorizable} & \textbf{> Random Baseline} & \textbf{Predictable} & \textbf{Emergent} \\
\hline
Causal judgement        & Functional   & 0  & No   & - & No \\
English Proverbs        & Functional   & 0  & No   & - & No \\
Implicatures            & Functional   & 0  & Yes  & Yes & No \\
NS words grammar        & Formal       & 38 & Yes  & No  & Yes \\
Rhyming                 & Formal       & 50 & No   & - & No \\
Tracking shuffled obj. & Functional   & 0  & No   & - & No \\
Commonsense QA          & Functional   & 3  & Yes  & Yes & No \\
GSM8K                   & Functional   & 0  & No   & - & No \\ \midrule
Analytic entailment     & Functional   & 4  & Yes  & Yes & No \\
Codenames               & Functional   & 0  & No   & - & No \\
Common morpheme         & Formal       & 0  & Yes  & Yes & No \\
Fact checker            & Functional   & 50 & Yes  & Yes & No \\
FoS detection           & Functional   & 0  & No   & - & No \\
Hindu knowledge         & Functional   & 50 & Yes  & No  & Yes \\
Logical deduction       & Functional   & 0  & No   & - & No \\
Misconceptions          & Functional   & 50 & Yes* & No  & Yes \\
Mod. arithmetic         & Functional   & 0  & No   & - & No \\
Phrase relatedness      & Functional   & 50 & Yes  & Yes & No \\
Physical intuition      & Functional   & 50 & Yes  & Yes & No \\
Social IQA              & Functional   & 0  & Yes  & Yes & No \\
Strange stories         & Functional   & 0  & Yes  & Yes & No \\
Strategy QA             & Functional   & 27 & Yes* & No  & Yes \\
\bottomrule
\end{tabular}
\caption{\label{tab:default-results-ntz} An overview of the tasks and a categorisation as formal or functional (\textbf{Competence Type}). The first 8 tasks are not previously identified to be emergent. For each task, we manually determine how many of 50 examples can be solved through memorisation (\textbf{Memorisable}). For a task to be \textbf{Emergent}, models must perform above the baseline (\textbf{> Random Baseline}) and the performance of the larger models must not be predictable based on that of smaller models (\textbf{Predictable}). This table is based on the zero-shot performance of the non-instruction-tuned 175B GPT-3 model \texttt{davinci}. * indicates that the increase above the random baseline is less than 5\%.}
\end{table*}

\subsection{Experimental Integrity and Generalisability}
\label{section:othermodels} 

To validate our experimental framework, particularly the use of BERTScore accuracy and our modifications to prompts, we conduct validity tests. 
These involve the evaluation of instruction-tuned models with in-context examples included in the prompts, referred to as the \emph{few-shot} setting, thereby enabling ICL in line with the experimental designs of prior work. The results of these tests replicated previous findings, confirming that our experimental framework does not hinder the potential for detecting emergent abilities. 

Since our findings rely on the use of LLMs that have not been instruction-tuned, we verify that the observed lower performance on tasks does not stem from the automatic metric (BERTScore) failing to evaluate model responses adequately. Specifically, if the model generates an answer that is correct, but does not align with the correct target option, BERTScore accuracy might fail to provide a reliable assessment. To this end, we conducted a post-hoc analysis by manually examining a subset of 50 outputs of non-instruction-tuned models from each task. Our focus was identifying instances where BERTScore accuracy failed to recognise correct responses (false negatives). Notice that false positives would not lead to an underestimation of model performance, and so have a lesser impact on our ability to identify emergence.
%The comprehensive details of our analysis, including our manual annotation of model outputs for each task, are included in the supplementary material accompanying this work. 
A comprehensive description of the analysis is included in Appendix \ref{sec:manualeval}. Our findings reinforce the notion that limitations -- inherent to all automatic evaluation -- do not detract from the overall validity of our results. 

Similarly, we perform other checks for potential aspects of our experimental setup that could lead to confounding effects in our results. These include manual analysis of model outputs to ensure the our prompts were interpreted correctly (Appendix \ref{sec:manualeval}), and the use of shortened outputs to enable easier evaluation (Appendix \ref{sec:shorteval}).

Finally, to ensure generalisability of our results, we extend our analysis to the LLaMA, Falcon, and T5 model families. Across each of these cases, a consistent pattern emerges: either task performance is predictable based on smaller model performance, or the performance is below the baseline. Overall, our analysis indicates that our experimental settings do not adversely affect our capacity to identify emergent abilities and our findings are generalisable across various model families.

\section{Instruction-Tuning as Implicit In-Context Learning}
\label{sec:t5performance}
The remarkable performance of instruction-tuned models cannot be solely attributed to their pre-training objective, which is to predict the next most probable token. This observation has led to the conjecture that models gain emergent functional linguistic abilities, such as reasoning~\cite{wei2022chain}. Nevertheless, LLMs exhibit several limitations that are at odds with this view: namely, their known sensitivity to minor prompt variations and their tendency to hallucinate. This leads us to hypothesise that the primary mechanism underlying the capabilities of instruction-tuned models may in fact be an indirect form of ICL, which we call `implicit in-context learning'. This section presents experimental results aimed at discerning whether this is the more plausible explanation underlying the performance of instruction-tuned LLMs.

Our evaluation in this section focuses on task solvability rather than performance. This is because the (sometimes wide) variation in parameter counts, architectures, and the pre-training data of the models we compare would necessarily mean that performance may differ across models. However, assessing task solvability offers a clearer insight into emergent abilities within the models. We utilise the previously-introduced BERTScore accuracy for all scenarios and evaluate models across the same 22 selected tasks previously outlined in Table \ref{tab:default-results-ntz}. In this setup, unlike the previous one, we only make use of non-instruction-tuned models in the setting wherein we provide examples in-context (few-shot), thereby eliminating concerns about the models' comprehension of task requirements.

\subsection{Comparative Analysis of Initial Tasks}
In discerning the more plausible explanation underlying the performance of instruction-tuned LLMs, our experiments are designed to yield differing outcomes based on whether models exhibit functional linguistic abilities or rely predominantly on ICL.

\begin{figure*}[ht!] 
\centering
\includegraphics[width=.95\textwidth]{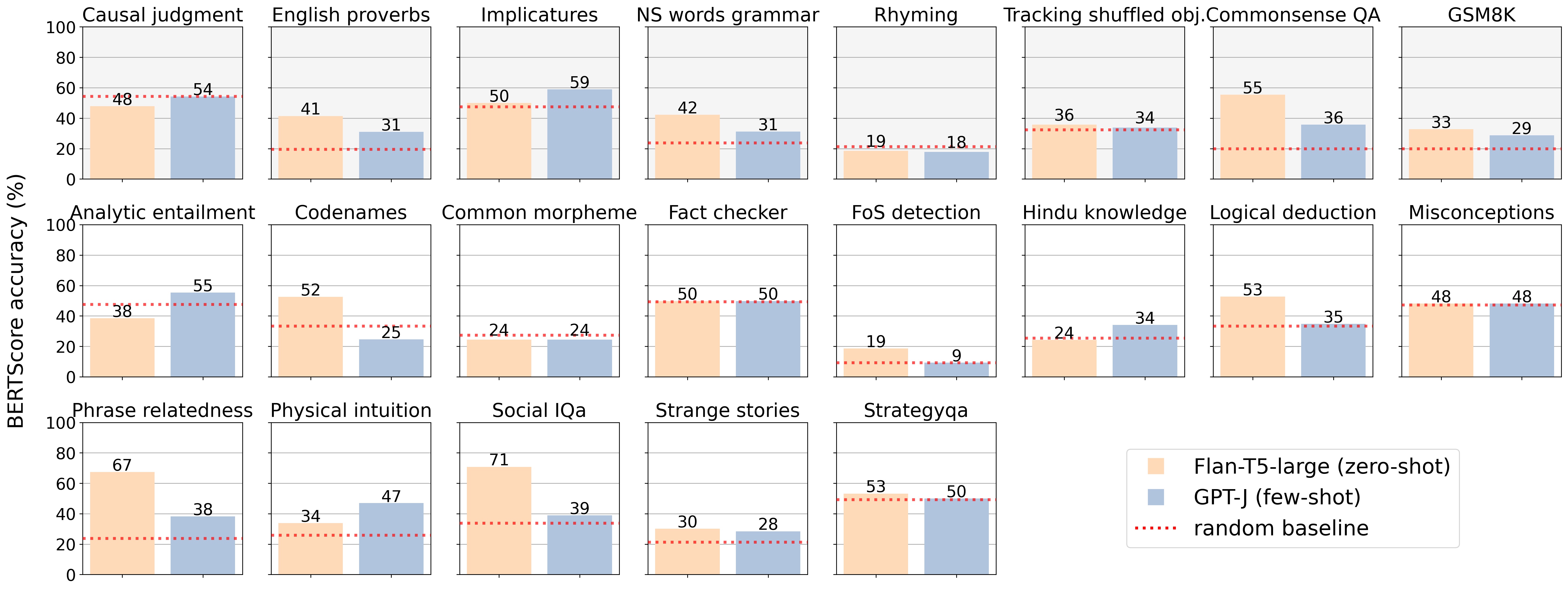}
\caption{\label{fig:t5-gpt-small} %A comparison of the tasks on which Flan-T5 performs above the baseline, and those on which the non-instruction-tuned (Non-IT) version of GPT 6.7B performs above baseline. 
The substantial overlap of the tasks on which the two models perform above the random baseline is noteworthy and indicates that instruction-tuning allows for the effective access of in-context capabilities rather than leading to the emergence of functional linguistic abilities. See text for details.
}
\end{figure*}

Specifically, we draw a comparison between the tasks that GPT-J (non-instruction-tuned, 6.7B) can successfully address in the few-shot setting, and those that can be solved by Flan-T5-large (instruction-tuned, 770M) in the zero-shot setting. The choice of these models is also based on the observation that there is no change in the model's performance between the zero-shot and few-shot settings for Flan-T5-large, indicating that it is too small for explicit ICL. On the other hand, we observe that there is a boost in performance across tasks in the few-shot setting for GPT-J, which indicates that it is capable of ICL. Notice that our choice of models ensures that the model we use to test which tasks can be solved using ICL is not instruction-tuned, and the model which is instruction-tuned is tested without in-context examples and also cannot explicitly access ICL. If instruction-tuning leads to models being capable of something fundamentally different from ICL  (for example, functional linguistic abilities), this would result in no substantial overlap in the set of tasks solvable solely through instruction-tuning and the set of tasks addressable solely via ICL. This comparison is presented in Figure \ref{fig:t5-gpt-small}. We exclude Modified arithmetic from this analysis, as the task is constructed in a manner that requires the use of in-context demonstrations.

Note the substantive dissimilarity between the two models we use: Flan-T5-large is an encoder-decoder model and GPT-J is a decoder only model. Additionally, they are trained on very different pre-training datasets, one is instruction-tuned while the other isn't, and they have very different parameter counts. Despite these fundamental differences, there is a substantial overlap in both the tasks where the two models exhibit above-baseline performance, as well as an overlap in the performance scores themselves. This overlap in the results underscores a compelling argument -- it is more likely that instruction-tuning serves as a mechanism that enables models to harness in-context capabilities more effectively, rather than the models having emergent reasoning abilities. There are exactly five of the 21 tasks we test wherein one model performs markedly above the random baseline while the other does not. Indeed, some of the cases are expected: in the case of Hindu knowledge, which is a recall-based task, GPT-J, which is larger than Flan-T5-large, has an advantage and performs better. Similarly, the highly instructional nature of the Codenames renders it particularly challenging fornon-instruction-tuned models. Of the remaining three tasks, the better-performing GPT-J only achieves an improvement of 5\% on Analytical entailment, which is binary classification. This leaves us with just Logical deduction, where Flan-T5-large benefits to some extent from the instructional nature of the questions, and Implicatures, where GPT-J achieves an accuracy of 59\%.

\subsection{Generalisability}
\label{ssec:instructcompare}

To evaluate if our results generalise to a further increase in model size and instruction-tuning data, we compare the tasks that can be effectively tackled by Flan-T5-large with those by instruction-tuned versions of the largest GPT models, i.e., \texttt{text-davinci-001} and \texttt{text-davinci-003} (additionally trained extensively on program code). It is important to note that these models have more than \emph{200 times} the number of parameters present in Flan-T5-large. We perform this comparison in the zero-shot setting, thus allowing us to compare the instruction-following capabilities of these models without triggering their ICL capabilities, which we know to increase markedly with scale. 

This comparison allows us to answer the following questions: a) Does increased scale largely impact the tasks on which models can perform above the random baseline, and b) Does enhanced instruction-tuning, including the incorporation of program code as seen in \texttt{text-davinci-003}, provide an advantage in being able to perform above the baseline on tasks? By limiting ourselves to the zero-shot setting, we ensure that our results are not affected by in-context capabilities, which we know to increase significantly with scale. 
Our results indicate that neither scale nor the inclusion of program code in instruction-tuning markedly alters the task solvability of a model. There is a substantial overlap in the tasks on which Flan-T5-large performs above the baseline and those on which \texttt{text-davinci-001} and \texttt{text-davinci-003} do: 16 of the 22 tasks we experiment with show this congruence. This overlap, and in several instances comparable performance across these diverse models, suggests that the effectiveness of instruction-tuning is consistent regardless of model scale or the nature of tuning datasets, in the absence of explicit ICL. Among non-overlapping tasks, certain recall-based tasks %(e.g., `Hindu knowledge') 
are better handled by larger GPT models due to their better recall abilities. %, aligning with earlier observations of their enhanced recall abilities. 
These results, illustrated in Figure \ref{fig:t5-gpt-bar}, Appendix \ref{app-bigplot}, confirm that our hypothesis, namely that `implicit in-context learning' is likely the primary mechanism in instruction-tuned LLMs, and that it is generalisable across model sizes and various instruction-tuning datasets. This also suggests that further scaling will probably not alter this trend.

\subsection{A Novel Theoretical Foundation}
Based on our observations on the capabilities and limitations of LLMs, we propose a novel alternative theory explaining why instruction-tuning helps models perform better: we propose that instruction-tuning enables models to map instructions to the form required for ICL, thus allowing instruction-tuned models to solve tasks using some implicit form of ICL. Importantly, during this process, models could be directly making use of the same underlying mechanism that makes ICL possible, just in a different way than when the model explicitly makes use of ICL from examples provided in the prompt. We call this use of ICL `implicit' in-context learning. Performing such a mapping would be relatively straightforward for a very large model, %particularly when the model is instruction-tuned. In fact, 
especially given that this task format aligns closely with the training process carried out during instruction-tuning. Investigating the exact nature of this mechanism is left for future work.

\section{Related Work}
\label{sec:emergent-explanation-problems}

\paragraph{Emergent Abilities} An \textit{emergent ability} was first defined as an ability that is not present in smaller models but is present in larger models \cite{wei2022emergent}. From a review of prior literature of LLMs including GPT-3, PaLM~\cite{DBLP:journals/jmlr/ChowdheryNDBMRBCSGSSTMRBTSPRDHPBAI23}, Chinchilla~\cite{hoffmann2022an}, Gopher~\cite{DBLP:journals/corr/abs-2112-11446} and LaMDA~\cite{DBLP:journals/corr/abs-2201-08239}, \citet{wei2022emergent} identified a total of 67 emergent abilities based on above-random performance of LLMs on tasks designed to test those abilities from the BIG-bench dataset~\cite{srivastava2023beyond}, and the Massive Multitask Language Understanding Benchmark~\cite{DBLP:journals/corr/abs-2009-03300}. Subsequent studies have explored additional abilities emergent in LLMs, such as Theory of Mind~\cite{kosinski2023theory} and cognitive biases \cite{Itzhak2023cognitive}.  
%Another work\cite{DBLP:journals/corr/abs-2304-10513} found that through the use of counterfactual tasks, performance of LLMs degrades in adversarial settings, which indicates that although LLMs exhibit some level of proficiency in abstract task-solving, they frequently lean towards employing context-specific approaches. 
However, ~\citet{schaeffer2023emergent} have previously questioned the existence of emergent abilities, arguing that emergence is likely to be a consequence of the discrete evaluation metrics commonly employed for assessing LLMs. Some~\cite{wei2022emergent} argue against this by pointing out that there are tasks on which LLMs are able to perform well above the random baseline where smaller models can only perform below it, suggesting that these abilities are still emergent and not just a consequence of discrete evaluation metrics. Similarly, several works~\cite{DBLP:journals/corr/abs-2304-11158, tefnik2023can, DBLP:journals/corr/abs-2307-02477, DBLP:journals/corr/abs-2304-10513} have explored the extent to which memory plays a role in LLMs' abilities.

\paragraph{In-Context Learning} ICL is a learning paradigm that has gained great popularity with the advent of LLMs \cite{DBLP:conf/nips/BrownMRSKDNSSAA20,DBLP:journals/csur/LiuYFJHN23}. ICL typically involves prompting an LLM with in-context demonstrations, and offers a more interpretable interface as well as greater computational efficiency compared to previous learning approaches \cite{dong2022survey,zhou2022large}. Notably, ICL has demonstrated strong performance on various natural language tasks \cite{kojima2022large,lampinen-etal-2022-language,wei2023larger}.

In terms of the theoretical rationale for ICL in LLMs, recent work indicates that it might share similarities with fine-tuning, in that it might allow models to ``learn'' from the examples presented in their prompt~\cite{dai2023why}. Similarly, it has been shown that ICL implements gradient descent implicitly and constructs a function at inference time on regression problems \cite{DBLP:conf/iclr/AkyurekSA0Z23,li2023transformers,DBLP:journals/corr/abs-2306-09927}, which may be related to gradient-based meta-learning \cite{DBLP:conf/icml/OswaldNRSMZV23}. A line of work shows that ICL is driven by the distributions of the pre-training data \cite{chan2022data,hahn2023theory}. Some other theoretical explorations attempt to explain ICL in terms of Bayesian inference \cite{xie2022an,li2023transformers,zhang2023and}.

To the best of our knowledge, none of the previous evaluations of emergent abilities have been conducted in a manner that explicitly distinguished between the ICL and instruction-tuning settings and prompting in the setting wherein these abilities are not triggered.% (i.e., zero-shot evaluation of non-instruction-tuned models). 

\section{Conclusions and Implications}
We started with two hypotheses: a) That the emergence of all previously-observed functional linguistic abilities is a consequence of ICL, and b) That the abilities which present themselves in instruction-tuned LLMs is more likely to be indicative of instruction-tuning resulting in implicit ICL, rather than the emergence of functional linguistic abilities. Our results confirmed both of these hypotheses.

The distinction between the ability to follow instructions and the inherent ability to solve a problem is a subtle but important one, and bears significance to the methods employed in utilising LLMs and the problems they are tasked with solving. Simple following of instructions without applying reasoning abilities produces output that is consistent with the instructions, but might not make sense on a logical or commonsense basis. This is reflected in the well-known phenomenon of `hallucination', in which an LLM produces fluent, but factually incorrect output \cite{DBLP:journals/corr/abs-2302-04023,doi:10.1126/science.adg7879} %, and can be mitigated using detailed instructions and examples.% \cite{DBLP:journals/corr/abs-2302-04023,shen2023chatgpt,doi:10.1126/science.adg7879}. 
The ability to follow instructions does not imply having reasoning abilities, and more importantly, it does not imply the possibility of latent, potentially-dangerous abilities. %~\cite{DBLP:journals/ais/Hoffmann23}. %Similarly, despite their zero-shot capabilities, in the absence of explicit examples (or in the case where the instructions weren't explicit enough), our framework would suggest that models will struggle. %This is in line with prior findings~\cite{mishra-etal-2022-reframing} which have shown that prompt engineering, as well as instruction and example design, are crucial for optimal LLM performance. 
Additionally, these observations imply that our findings hold true for any model which exhibits a propensity for hallucination or requires prompt engineering, including those with greater complexity, regardless of scale or number of modalities, such as GPT-4. %Finally, we note that our results are directed to those safety concerns arising out of latent hazardous abilities and do not extend to other, user-directed safety concerns resulting from the use of LLMs by malicious actors. %However, we note that these concerns, unlike that of latent hazardous abilities, are not limited to LLMs alone.
%Our results point to a need for a more thorough analysis of tasks along the lines of memorisability, data leakage, quality (e.g., number of examples in the test set), and a classification of such tasks into those requiring formal linguistic abilities and functional linguistic abilities~\cite{DBLP:journals/corr/abs-2301-06627}.
By contributing to a deeper understanding of these models' abilities and limitations, we help to demystify LLMs, alleviate their related safety concerns, and lay out a framework for their more efficient use. 
%Ultimately, this absence of evidence for the emergence of functional linguistic abilities represents a significant step towards instilling trust in language models and leveraging their abilities with confidence, since it is indicative of the complete lack of latent hazardous abilities in LLMs, as well as of their user-controllability. 

\section*{Limitations}
Although we experiment on an extensive amount of model sizes across various architectures (e.g., T5, GPT, Falcon, LLaMA), we were unable to ensure an exact match of parameter counts across the different architectures. This is due to the variation in the publicly-available releases of these models. In this work, we used all models at the parameter counts that were available. However, another alternative would be to conduct pre-training to ensure equal parameter counts and comparable pre-training data, though this would involve a substantial computational investment. In all tasks, there is a risk of data leakage, especially for LLMs whose training datasets are not publicly known. In this work, we assume that data leakage has not occurred beyond what was reported in official publications for specific models (e.g., BIG-bench for GPT-4). As such, we do not consider data leakage a factor when we consider a task to be `memory-based', although, in practice, the presence of data leakage can have a biasing effect on model performance. Our experiments are limited to English tasks. This is primarily a consequence of previous work on emergent abilities and on the limitations of computational budget to run experiments on other languages. We intend to focus future work on datasets that include other languages including low resource languages. 

\section*{Ethical Considerations}
Our work does not imply that LLMs have absolutely no potential for harm. By leveraging the sophisticated linguistic capabilities of LLMs, malicious actors can craft highly convincing and personalised fake news articles or phishing messages, which may become increasingly difficult to distinguish from legitimate messages. The ease and efficiency with which LLMs can be used for these purposes highlight the need for detection mechanisms, along with ethical guidelines to mitigate the risks and protect individuals and democratic processes. Similarly, identifying that LLM capabilities are not a precursor to an AI-driven existential threat does not eliminate the need for ongoing vigilance in AI safety research. Our findings present an unique opportunity to prioritise the most pressing aspects of LLM safety while simultaneously exploring research avenues beyond mere scaling up.

We recognise that the conversation about LLMs' capabilities and limitations plays a crucial role in the broader social discourse on AI. This underscores the importance of thoughtful consideration and a high degree of care in all related research and publication efforts.

\section*{Acknowledgements}
This work has been funded by the LOEWE Distinguished Chair “Ubiquitous Knowledge Processing”, LOEWE initiative, Hesse, Germany (Grant Number: LOEWE/4a//519/05/00.002(0002)/81). This research work has been funded by the German Federal Ministry of Education and Research and the Hessian Ministry of Higher Education, Research, Science and the Arts within their joint support of the National Research Center for Applied Cybersecurity ATHENE. We would also like to thank the Early Career Research grant from the University of Bath. This work would not have been possible without the generous grant from the Microsoft Accelerate Foundation Models Academic Research fund, which allowed us to experiment extensively with the Azure OpenAI service.

% \section*{Author Contributions}
% Sheng Lu, Irina Bigoulaeva, and Rachneet Sachdeva conducted the experiments. Harish Tayyar Madabushi conceived the experiments and analysed the results. Iryna Gurevych supervised this work. Harish Tayyar Madabushi and Irina Bigoulaeva wrote the manuscript and all authors reviewed the manuscript. 

% \noindent \textsuperscript{*}Sheng Lu and Irina Bigoulaeva are equal first authors.

% Entries for the entire Anthology, followed by custom entries
\bibliography{custom}
\newpage
\onecolumn

\addtocontents{toc}{\protect\setcounter{tocdepth}{0}}

\appendix

\section{In-Context Learning and Instruction-Tuning}
\label{app:iclit-fig}

\begin{figure*}[htp!]
\centering
\includegraphics[width=.98\textwidth]{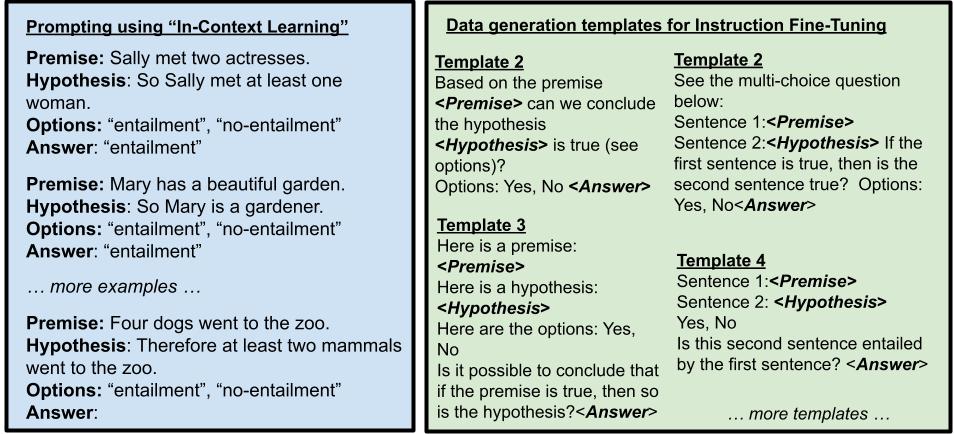}
\caption{\label{fig:instructTuned} The figure on the left depicts prompting using ICL, where the model infers the task and the patterns based on a few examples. The figure on the right presents a few of the templates used to generate instruction fine-tuning data which models are fine-tuned on to allow them to better interpret prompts. The task depicted in these examples is Analytical entailment and the templates are from the Flan instruction fine-tuning dataset~\cite{wei2022finetuned}.}
\end{figure*}

\section{Controls for Possible Bias}
\label{sec:bias}
 In order to ensure that our evaluation is fair, we identify potential biases that could influence our findings and design our experiments to mitigate such biases. In cases where this is not possible, we shape our experiments to maximise our chances of identifying emergent abilities, if they do indeed exist.

\subsection{Prompt Formats}
\label{sec:prompts}

 \begin{table}[ht!]
 \footnotesize
     \centering
     \begin{tabular}{p{0.35\linewidth}p{0.5\linewidth}}
     \toprule
          \textbf{Prompt format} & \textbf{Example} \\ \midrule
          \multirow{6}{\linewidth}{\texttt{default}, closed} & \textbf{Question:} Austin's family was celebrating their parents 50th anniversary during dinner at a new restaurant. What would Austin's family do next? From the following choices, choose the correct answer: ``Refuse to eat dinner with the family'', ``Eat dinner at the restaurant'', ``Happy'' \newline \textbf{Answer:} \\ \midrule
          \multirow{3}{\linewidth}{\texttt{completion}, open} & Austin's family was celebrating their parents 50th anniversary during dinner at a new restaurant. What would Austin's family do next? The correct answer is \\ \hline
          \multirow{5}{\linewidth}{\texttt{completion}, closed} & Austin's family was celebrating their parents 50th anniversary during dinner at a new restaurant. What would Austin's family do next? The possible answers are ``Refuse to eat dinner with the family'', ``Happy'', ``Eat dinner at the restaurant'', but the correct answer is \\ \midrule
          \multirow{6}{\linewidth}{\texttt{adversarial}, closed} & \textbf{Question:} Austin's family was celebrating their parents 50th anniversary during dinner at a new restaurant. What would Austin's family do next? \newline \textbf{Options:} (a) ``Refuse to eat dinner with the family'', (b) ``Eat dinner at the restaurant'', (c) ``Happy'' \newline \textbf{Answer:} \\
          \bottomrule
     \end{tabular}
     \caption{Sample prompts of the three formats we use. The samples are from the Social IQA task of BIG-bench.}
     \label{tab:prompt_formats}
 \end{table}

 Table \ref{tab:prompt_formats} shows an example of each of our prompt formats. We make two important changes to the prompting strategies used: First we refine all prompts to ensure their solvability even in the absence of instruction comprehension. We call this adjusted prompt format the \textit{completion-style prompt}, and use it for all models (See Table \ref{tab:prompt_formats}). We experiment with minor variations to these prompts so as to find the most optimal format.

 This change is necessary, since in order to assess the true abilities of non-instruction-tuned models in the zero-shot setting, it is imperative to evaluate their ability to accurately perform tasks without relying on explicit instructions. Many of the tasks presented in Section \ref{sec:tasks} (Tasks) involve prompts that inherently require an understanding of explicit instructions. Since LLMs in their base form are trained to perform next-word prediction, it is unreasonable to expect that without instruction-tuning, they will respond adequately to multiple choice question prompts requiring them to pick the correct answer from a set of options. We hypothesised that using such a prompt style would give an unequal advantage to the instruction-tuned models. Indeed, our initial prompt experiments demonstrated that non-instruction-tuned models merely try to ``complete'' the text of the prompt by generating additional answer choices, sometimes even additional new questions. However, once the prompt itself was adjusted to take the form of a sentence to be completed, non-instruction-tuned models were likelier to output one of the answer choices. We confirm that these changes do not skew our results by replicating prior results using instruction-tuned models, which we use as a baseline. 

 The second change we make to our prompting strategy involves the exploration of two types of completion-style prompts: \textit{closed} and \textit{open}. In the closed prompt format, we provide answer choices alongside the prompt, while in the open prompt format, the answer choices are withheld. We find that when models are prompted using the open prompt strategy, their generated results often exhibit little or no resemblance to the provided answer choices. Consequently, evaluating the correctness of the generated answers becomes challenging. As a result, experiments utilising the open prompt setting are completely excluded from our analysis. However, we provide access to these responses in the data accompanying this study, allowing other researchers to experiment with it.

\subsection{Validation through Shortened output Generation}
\label{sec:shorteval}
\begin{figure*}[]
\centering
\includegraphics[width=.85\textwidth]{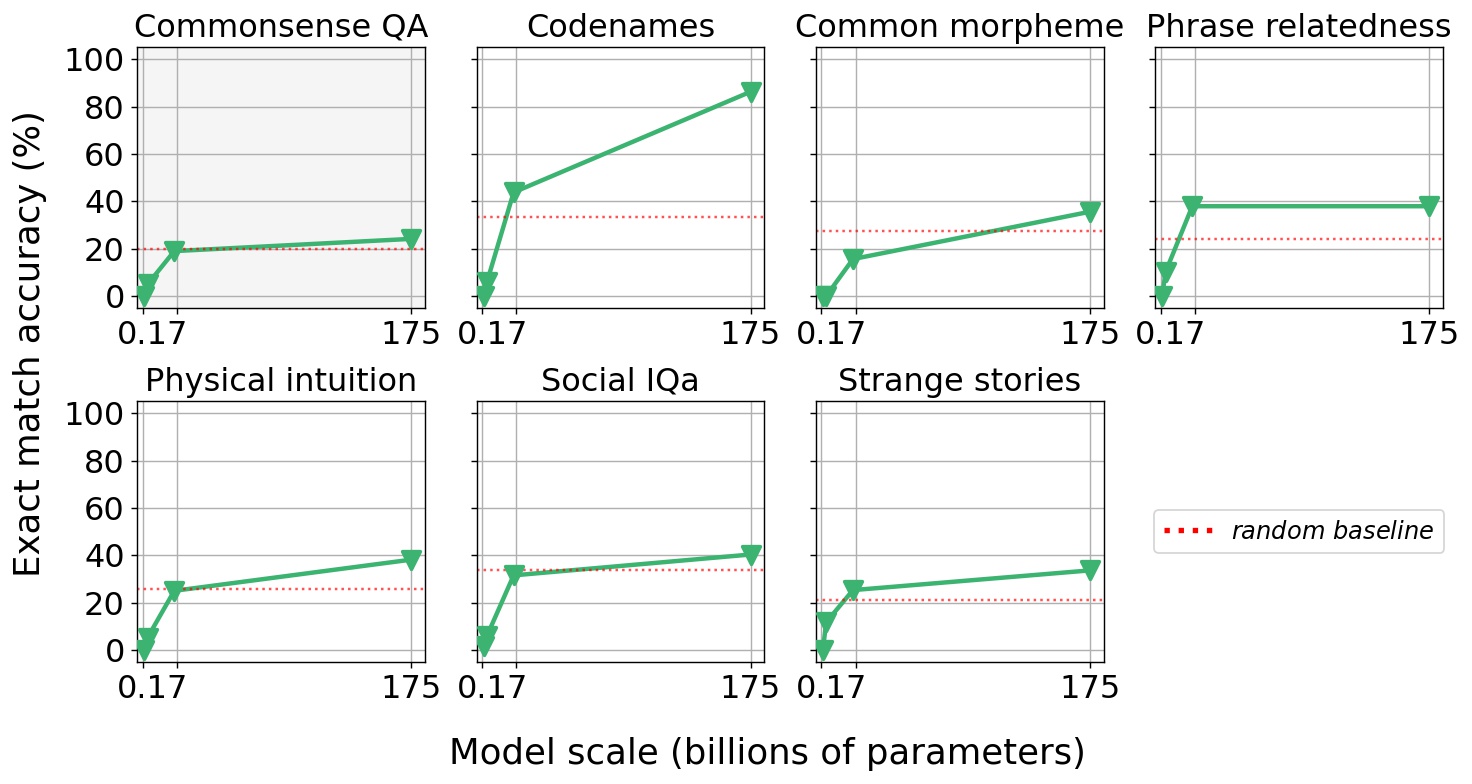}
\caption{\label{fig:adv-gpt-zero} Performance of non-instruction-tuned GPT models using the \texttt{adversarial} prompt on the subset of tasks wherein the performance is above the random baseline. The subplot with grey background indicates that the task is not previously identified to be emergent. The performance on Codenames, Phrase relatedness, and Strange stories is predictable and so not emergent. Across the remaining tasks, the improvements in performance compared to the random baseline are relatively modest. Additionally, of the tasks on which the performance gain is slightly more notable, we find that Physical intuition is a memory intensive task and Common morpheme has a small test set. }
\end{figure*}

LLMs lacking instruction-tuning often exhibit a degree of proficiency in adhering to instructions, albeit within constrained limits, particularly in the context of models with a substantial parameter count of 175B~\cite{wei2022finetuned}. We leverage this phenomenon by using the ``adversarial prompt setting'', wherein the model is required to generate output choices, such as options ``a'' or ``b'' instead of the target choice. In this setting we evaluate models using a relaxed version of exact match wherein an answer is marked correct if it contains the correct target option. This flexibility is once again designed to allow us to detect any possible indication of emergence. Note that this assessment allows us to circumvent the necessity for employing less precise evaluation criteria as is required when evaluating more verbose responses. The results of this evaluation on the seven of 22 tasks wherein the performance is above the random baseline are presented in Figure \ref{fig:adv-gpt-zero}. 

Of these seven tasks on which the non-instruction-tuned version of GPT-3 performs above the random baseline, three are predictable based on the performance of smaller models and thus not considered emergent. The only task on which the improvement over the baseline is not predictable and notable is `physical intuition.' This task includes questions such as ``The bonds in sodium chloride are of what type? Options: Ionic: 1, Covalent: 0, Metallic: 0, Hydrogen: 0'', which are likely to be more memory based. Common morpheme, on the other hand, is a non-trivial task that require `reasoning' abilities. However, we find that it has an extremely small test set with only fifty examples and thus the improvement in accuracy is only a small fraction of the total. As such, even in this setting, where we need not employ the less precise evaluation criteria, we find no evidence for the emergence of functional linguistic abilities.

\subsection{Manual Evaluation of Responses}
\label{sec:manualeval}
To ensure that our results are not biased, we present here a manual analysis of 50 output examples from each task, the results of which is presented in Table \ref{tab:manualeval}. Recall that modified arithmetic, GSM8K, and codenames are always evaluated using exact match accuracy and so are not included in this analysis.

\begin{table}[ht!]
\footnotesize
\centering
\begin{tabular}{p{0.42\linewidth}ccc}
\toprule
\textbf{Task}               & \textbf{BSA\%}  & \textbf{MA\%}  & \textbf{Base\%} \\ \hline
Analytic entailment	        & 48	 & 14    & 48 \\
Common morpheme	            & 32	 & 22    & 27 \\
English proverbs	        & 10	 & 6     & 20 \\
Fact checker	            & 52	 & 34    & 49 \\
Figure of speech detection	& 10	 & 10    & 9  \\
Hindu knowledge	            & 52	 & 54    & 25 \\
Implicatures	            & 58	 & 6     & 48 \\
Misconceptions	            & 48	 & 40    & 47 \\
Nonsense words grammar	    & 34	 & 22    & 24 \\
Phrase relatedness	        & 44	 & 34    & 24 \\
Physical intuition	        & 46	 & 40    & 26 \\
Rhyming	                    & 16	 & 6     & 21 \\
Social IQA	                & 36	 & 38    & 34 \\
Strategy QA	                & 58	 & 58    & 49 \\
Tracking shuffled objects	& 34	 & 20    & 32 \\
Strange stories	            & 34	 & 28    & 21 \\
\hdashline
Logical deduction	        & 26	 & 34    & 33 \\
Causal judgement	        & 46	 & 56    & 54 \\
\hdashline
Commonsense QA	            & 36	 & 54    & 20 \\
\bottomrule
\end{tabular}
\caption{\label{tab:manualeval} A comparison of BERTScore Accuracy (\textbf{BSA\%}) and a manual evaluation accuracy (\textbf{MA\%}) on 50 examples from each task. The analysis reveals that in instances of notable disparity, BERTScore accuracy generally tends to result in false positives (top block). In exactly three cases BERTScore accuracy underestimates performance: in two instances the increase allows model performance to increase above the baseline only marginally. In the case of Logical deduction, the model sometimes produces answers that are copied from the question but are still technically correct answers, which could lead to the MA\% score being too lenient. In the case of Causal judgement, the increase is only slight compared to the above 50\% baseline. Where there is a substantial performance boost above the baseline (bottom block), this particular task's predictability based on smaller model performance implies that it remains not emergent. As such, we find that even a lenient manual scoring does not affect our conclusion.}
\end{table}

In Table \ref{tab:manualeval}, `BERTScore accuracy \%' represents the percentage of correct answers as determined by the automatic metric of the 50 examples selected for manual evaluation and `manual evaluation accuracy \%' represents the percentage of correct answers as determined by a manual analysis of the results by one of the authors on the same set of examples. Recall that the purpose of this exercise is to ensure that the automatic evaluation metric does not affect our conclusion in terms of the existence of emergent abilities. Our analysis shows that, in the majority of cases, the automatic metric overestimates model performance. This set of tasks is represented in the top block in Table \ref{tab:manualeval}. 

In the case of Logical deduction, the model sometimes produces answers that are copied from the question but are still technically correct answers, which could lead to the MA\% score being too lenient. In the case of Casual judgement, the increase is only slight compared to the above 50\% baseline. These two cases wherein the manual evalution indicates a higher scores are represented in Table \ref{tab:manualeval} block 2. Finally, on `Commonsense QA', the only task where there is a marked increase over the baseline, such performance is predictable based on the performance of smaller models, and so the task is not emergent. 

This analysis of 50 examples from each task carries a degree of imprecision. Crucially, however, it is imperative to recognise that our primary objective is to ensure that these inaccuracies, inherent to the automatic evaluation of generative models, do not fundamentally alter our conclusions. Our analysis underscores that this is the case and that these limitations do not undermine the validity of our findings. 

Similarly, we study the output of non-instruction-tuned models to ensure that they are able to interpret the instructions in the questions. Our qualitative analysis points to them indeed being able to interpret task requirements. For example, in the `Causal judgement' task, models produce `yes' or `no' answers,  as required by the task.  Additionally, we note the above-baseline performance of the non-instruction-tuned models on \emph{some} tasks, albeit not functional linguistic tasks, which further lends support to the notion that such models have access to information pertaining to task requirements, once again confirming the validity of our findings.  

\section{Experimental Setup}
\label{app:experimental-setup}

{ % begin box to localize effect of arraystretch change
\renewcommand{\arraystretch}{1.2}
\footnotesize
\begin{table}[ht!]
\small
\centering
\begin{tabular}{lc}
\toprule
\textbf{Model}   & \textbf{Tasks} \\ \midrule
GPT-2            & \multirow{13}{*}{All of the 22 selected tasks} \\
GPT-2-IT         & \\
GPT-2-XL         & \\
GPT-2-XL-IT      & \\
GPT-J            & \\  
GPT-JT           & \\
\texttt{davinci} & \\
\texttt{text-davinci-001} & \\
\texttt{text-davinci-003} & \\
T5-small          & \\
Flan-T5-small     & \\
T5-large          & \\  
Flan-T5-large     & \\ \midrule
Falcon-7B         & \multirow{7}{*}{Logical deductions, Social IQA, GSM8K, Tracking shuffled objects} \\
Falcon-7B-Instruct      & \\
Falcon-40B        & \\
Falcon-40B-Instruct     & \\
LLaMA-7B          & \\
LLaMA-13B         & \\
LLaMA-30B         & \\
\bottomrule
\end{tabular}
\caption{\label{tab:experiments} An overview of the experimental setup. Models in the GPT and T5 families are evaluated on all tasks and those in the Falcon and LLaMA families on a subset of representative tasks. In addition, each evaluation is performed in the closed and closed adversarial prompting strategies.}
\end{table}
}

This section provides additional details of our experimental setup previously presented in Section \ref{sec:experiments-main}. As discussed, we evaluate each of the 12 models selected from the T5 and GPT families (Section \ref{sec:models}) on all of the 22 selected tasks, while those in the Falcon and LLaMA families are evaluated on a subset of representative tasks, namely: Logical Deductions, Social IQA, GSM8K, and Tracking Shuffled Objects.%as described in Section \ref{section:tasks}.
For each case, we employ the prompting strategies: closed, and closed adversarial, as discussed in Section \ref{sec:prompts}. In addition, we evaluate each model and prompting strategy using both the few-shot and the zero-shot settings. When using the few-shot setting, we use 5 in-context examples. To ensure reproducibility, we use the test sets provided by the tasks. Statistics associated with the test sets are included in the BIG-bench description~\footnote{\url{https://github.com/google/BIG-bench}}. To consider the variability in responses, we conduct each experiment three times and calculate the average result. All experiments that we run locally are run on NVIDIA A100 GPUs using a temperature of 0.01 and a batch size of 16. Our locally-run experiments took approximately between 8 and 12 hours per task, depending on the size of the test sets. In the case of GPT-3 175B parameter models (davinci, text-davinci-001, and text-davinci-003), we make use of the official API for evaluation which is done once using a temperature of 0 to aim for deterministic output. The total cost of our API usage was approximately \$1,500. While we restrict our evaluation to a single run due to cost constraints, it's improbable that this will impact the results of our experiments. This is because we also set the temperature to 0, which guarantees result reproducibility and minimises the possibility of hallucinations.

 In addition, we evaluate six selected models from the LLaMA and Falcon families (see Section \ref{sec:models}), on four of the 22 tasks chosen earlier. We pick these four tasks ensuring that two have been previously identified as emergent (Logical Deductions and Social IQA) and the other two have been determined to be non-emergent (GSM8K and Tracking Shuffled Objects). Once again we test these using the closed and adversarial prompting strategies and run each experiment thrice to account for variance. 
 Lastly, to avoid relying solely on discrete metrics for evaluating emergence, we employ four evaluation metrics: exact match, BERTScore accuracy, continuous BERTScore, and edit distance, as described in Section \ref{sec:evalmetrics}. 
 
In evaluating BERTScore accuracy, evaluate models based on the semantic similarity between the output text and the provided answer choices using BERTScore~\cite{DBLP:conf/iclr/ZhangKWWA20}\footnote{BERTScore V 0.3.13 using RoBERTa Large, 355M parameters, available at \url{https://huggingface.co/FacebookAI/roberta-large/commit/716877d372b884cad6d419d828bac6c85b3b18d9}} 

In terms of a random baseline, given the variable number of options associated with some of the tasks under evaluation, we construct the baseline for each task by randomly selecting options for questions in that task multiple times and finding an average score. 

\section{Additional Results: Implicit In-Context Learning}
\label{app-bigplot}
\begin{figure*}[ht!] 
\centering
\includegraphics[width=.9\textwidth]{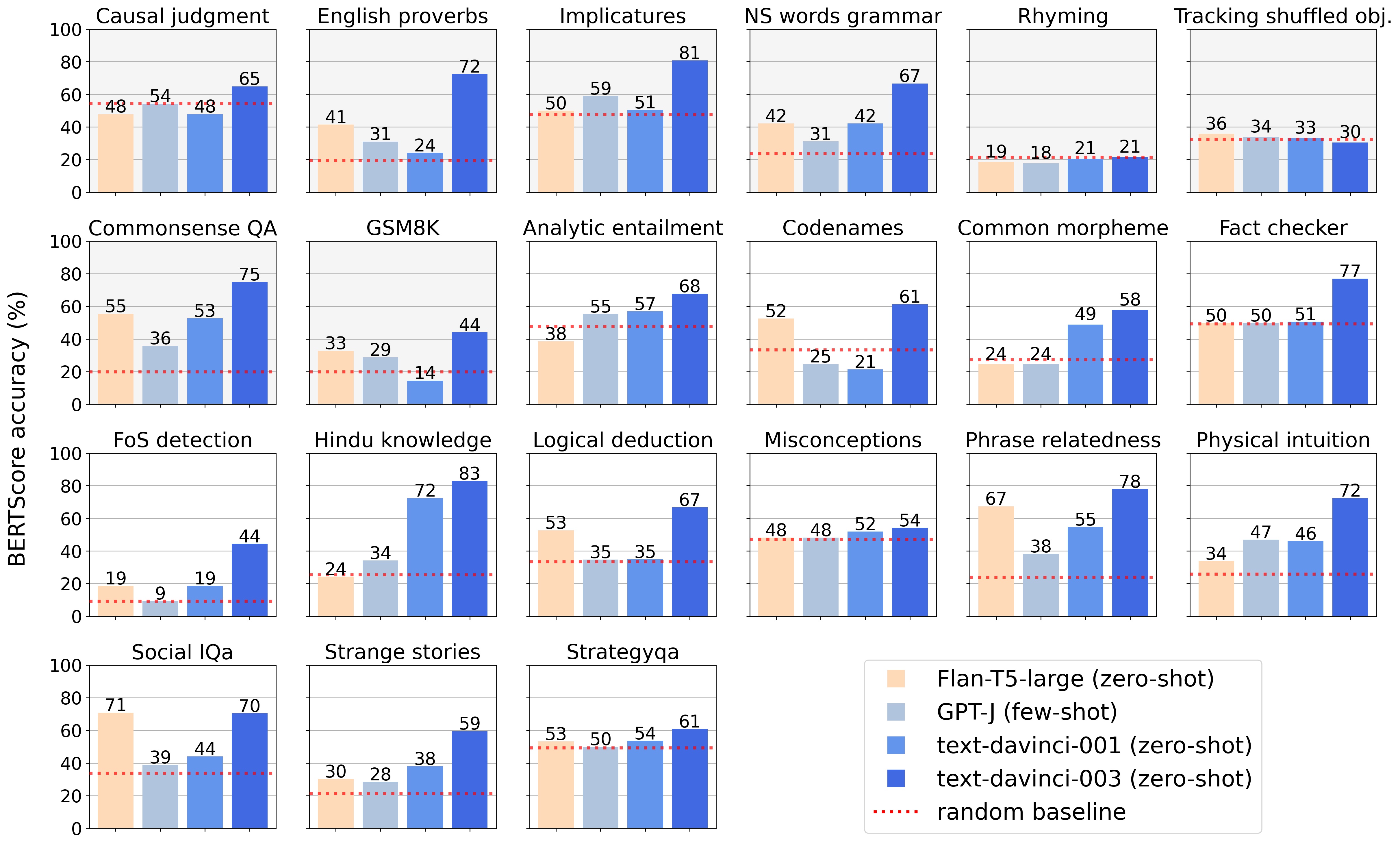}
\caption{\label{fig:t5-gpt-bar} %A comparison of the tasks on which Flan-T5 performs above the baseline, and those on which the non-instruction-tuned (Non-IT) version of GPT 6.7B performs above baseline. 
A comparison of the performance of Flan-T5-large (zero-shot), GPT-J (few-shot), \texttt{text-davinci-001} (zero-shot), and \texttt{text-davinci-003} (zero-shot) using the \texttt{completion} prompt. The subplots with grey background are results for tasks that are not previously identified to be emergent. Modified arithmetic is excluded from the analysis, as the task is constructed in a manner that requires the use of in-context demonstrations. The substantial overlap of the tasks on which the two models perform above the random baseline is noteworthy and indicates that instruction-tuning allows for the effective access of in-context capabilities rather than leading to the emergence of functional linguistic abilities.
}
\end{figure*}

\section{Detailed Task Information}
\label{app:tasks}

In this section, we give a detailed overview of our chosen tasks. For each task, we provide the task description and a selected example to illustrate the style of the questions and answers (Table \ref{tab:task_description} below). Our choice of tasks includes those tasks which were found to be emergent in GPT-3, primarily from BIG-bench. BIG-bench is licenced under the Apache-2.0 license, and our use of the dataset, based on the license and the description of provided is consistent with its intended use. This dataset contains no personally identifiable data and is designed to evaluate a range of reasoning and linguistic abilities in LLMs. 
% \begin{table*}[]
% \small
% \centering
%     \begin{tabular}{lll}
%     \hline
%     Cognitive Skill & Subcategory & Task Name \\
%     \hline
%         \multirow{1}{10em}{Formal Competence} & \multirow{1}{10em}{Linguistic Knowledge} & code names \\
%         & & common morpheme \\
%         & & nonsense words grammar \\
%         & & phrase relatedness \\
%         & & rhyming \\
%         & & \\
%     \multirow{1}{10em}{Functional Competence} &
%    \multirow{1}{10em}{Formal Reasoning} & analytic entailment \\ 
%     & & gsm8k \\
%     & & logical deduction \\
%     & & modified arithmetic \\
%     & & tracking shuffled objects \\
%     & & \\
%     & \multirow{1}{10em}{World Knowledge} & fact checker \\ 
%     & & hindu knowledge \\
%     & & misconceptions \\
%     & & physical intuition \\
%     & & social iqa \\
%     & & strategyqa \\
%     & & \\
%     & \multirow{1}{10em}{Situation Modeling} & causal judgment \\
%     & & vitaminc fact verification \\
%     & & \\
%     & \multirow{1}{10em}{Communicative Intent} & english proverbs \\ 
%     & & figure of speech detection \\
%     & & implicatures \\
%     & & strange stories \\
%          \hline
%     \end{tabular}
%     \caption{Our chosen tasks categorized into the cognitive skills categories from \citet{DBLP:journals/corr/abs-2301-06627}. Categorization was done manually and mutually agreed upon by the authors.}
%     \label{tab:task_list}
% \end{table*}

\begin{small}
\begin{longtblr}[
   caption={List of our chosen tasks along with their brief description and sample inputs.},
   label={tab:task_description}
]{
  colspec={ @{} X[halign=l] X[1.7,halign=l] X[1.7,halign=l] @{} },
  row{1}={halign=l},
  rowhead = 1
}
\toprule
\textbf{Task Name} & \textbf{Description} & \textbf{Example} \\ \midrule
\SetCell{r=4,halign=l}{}Causal judgement & This task tests whether large language models can comprehend a short story that introduces multiple cause-effect events. & \textbf{Input:} The CEO of a company...Did the CEO intentionally harm the environment? \newline \textbf{Options:} Yes, No \newline \textbf{Target:} Yes \\ \midrule

\SetCell{r=4}{}English proverbs & This task asks models to find the English proverb corresponding to a given story. &  \textbf{Input:} Both Tim and John...Which of the following proverbs best apply to this situation? \newline \textbf{Options:} ``Ignorance is bliss'', ``A bad thing never dies''... \newline \textbf{Target:} Ignorance is bliss \\ \midrule

\SetCell{r=4}{}Implicatures & This task asks models to predict whether one speaker's answer to another counts as a yes or as a no. & \textbf{Input:} Speaker 1: ``But aren't you afraid?'' Speaker 2: ``Ma'am, sharks never attack anybody.'' \newline \textbf{Options:} Yes, No \newline \textbf{Target:} No \\ \midrule

\SetCell{r=4}{}Nonsense words grammar & This task requires the language model to guess the grammatical role of nonsense words. & \textbf{Input:} Which word in the following sentence is a verb? The grilshaws bolheavened whincely. \newline \textbf{Options:} The, grilshaws, bolheavened, whincely \newline \textbf{Target:} bolheavened \\ \midrule

\SetCell{r=3}{}Rhyming & This task measures how well language models can understand rhyming in English. &  \textbf{Input:} What rhymes with cruise? \newline \textbf{Options:} disaster, creates, disguise, listen, crews \newline \textbf{Target:} crews \\ \midrule

\SetCell{r=4}{}Tracking shuffled objects & This task tests a model's ability to work out the final state of a system given its initial state and a sequence of modifications. & \textbf{Input:} Alice, Bob, and Claire are playing a game...At the end of the game, Alice has the \newline \textbf{Options:} ``orange ball'', ``white ball'', ``blue ball'' \newline \textbf{Target:} blue ball \\ \midrule

\SetCell{r=4}{}Commonsense QA & This task requires the models to answer commonsense questions based on their rich prior knowledge. & \textbf{Input:} Sammy wanted to go to where the people were. Where might he go? \newline \textbf{Options:} ``race track'', ``populated areas''... \newline \textbf{Target:} populated areas \\ \midrule

\SetCell{r=4}{}GSM8K & The dataset supports the task of question answering on basic mathematical problems that require multi-step reasoning. & \textbf{Input:} Weng earns \$12...How much did she earn? \newline \textbf{Options:} 13, 17, 10, 8, 25 \newline \textbf{Target:} 10 \\ \midrule

\SetCell{r=4}{}Analytic entailment & This task presents pairs of sentences and tests a model's ability to determine whether there is linguistic entailment. & \textbf{Input:}  Lina met two nurses. So, Lina met at least one woman. \newline \textbf{Options:} entailment, no\_entailment \newline \textbf{Target:} no\_entailment \\ \midrule

\SetCell{r=4}{}Codenames & This task asks models to identify words associated with a given word. & \textbf{Input:} Try to identify the 3 words best associated with the word INFRASTRUCTURE from the following list: ant, genie, government, cable... \newline \textbf{Target:} ant, cable, government \\ \midrule

\SetCell{r=4}{}Common morpheme & This task ask models to select the most likely option for the meaning of the morpheme that is common among an input list of words. & \textbf{Input:} What is the common morpheme among these words: pyre, empyrean, antipyretic, pyrotechnics \newline \textbf{Options:} fire, hot, oxygen, medicine \newline \textbf{Target:} fire \\ \midrule

\SetCell{r=4}{}Fact checker & This task tests models' ability to evaluate claims as true or false. & \textbf{Input:} On June 2017, the following claim was made... \newline \textbf{Question:} Was this claim true or false? \newline \textbf{Options:} true, false \newline \textbf{Target:} true \\ \midrule

\SetCell{r=5}{}Figure of speech detection & This task asks a model to detect which figure of speech is embodied by each of the example English sentences/phrases shown. & \textbf{Input:} Please identify the figure of speech embodied by the following English sentences. \newline \textbf{Sentence:} They fought like cats and dogs. \newline \textbf{Options:} Simile, Metaphor... \newline \textbf{Target:} Simile \\ \midrule

\SetCell{r=4}{}Hindu knowledge & This task asks models to answer questions about Hindu mythology. & \textbf{Input:} In Hinduism, the principle deity associated with creation is whom? \newline \textbf{Options:} Brahma, Shiva, Rama, Vishnu \newline \textbf{Target:} Brahma \\ \midrule

\SetCell{r=3}{}Logical deduction & This task requires deducing the order of a sequence of objects from a minimal set of conditions. & \textbf{Input:} On a shelf, there are three books... \newline \textbf{Options:} ``The black book is the leftmost''... \newline \textbf{Target:} The black book is the leftmost \\ \midrule

\SetCell{r=3}{}Misconceptions & This task measures whether a model can discern popular misconceptions from the truth. & \textbf{Input:} Twinkies are edible for decades or longer. \newline \textbf{Options:} T, F \newline \textbf{Target:} F \\ \midrule

\SetCell{r=5}{}Modified arithmetic & This task asks a model to perform a mathematical operation. & \textbf{Input:} In the following lines, the symbol -> represents a simple mathematical operation.\newline102 + 435 -> 537...466 + 214 -> \newline \textbf{Options:} 672, 680, 686 \newline \textbf{Target:} 680 \\ \midrule

\SetCell{r=5}{}Phrase relatedness & This task presents models with a phrase (n-gram), and asks them to select the most related phrase (n-gram) among the choices. & \textbf{Input:} For each word or phrase, identify the most related choice from the listed options. \newline home town \newline \textbf{Options:} ``location'', ``native city''... \newline \textbf{Target:} native city \\ \midrule

\SetCell{r=4}{}Physical intuition & This task asks models to deduce the physical mechanism or behavior associated with a physical system. & \textbf{Input:} A bug hits the windshield of a car. Does the bug or the car accelerate more due to the impact? \newline \textbf{Options:} Bug, Car, Neither \newline \textbf{Target:} Bug \\ \midrule

\SetCell{r=4}{}Social IQA & This task measures the ability of models to reason about the common-sense implications of social situations. & \textbf{Input:} Tracy didn't go home that evening and resisted Riley's attacks. What does Tracy need to do before this? \newline \textbf{Target:} ``Make a new plan'', ``Find somewhere to go''... \newline \textbf{Target:} Find somewhere to go \\ \midrule

\SetCell{r=5}{}Strange stories & This task measures the emotional intelligence of language models through a psychology test with naturalistic short stories. & \textbf{Input:} At school today... \newline \textbf{Question:} How would Ben's mom feel if she later learned that John was not at school? \newline \textbf{Options:} worried, confused, fearful, joyful \newline \textbf{Target:} confused \\ \midrule

\SetCell{r=5}{}Strategy QA & This is a question-answering benchmark focusing on open-domain questions where the required reasoning steps are implicit in the question and should be inferred using a strategy. & \textbf{Input:} Is it common to see frost during some college commencements? \newline \textbf{Options:} Yes, No \newline \textbf{Target:} Yes \\
\bottomrule
\end{longtblr}
\end{small}

\newpage

\section{Task Memorisability}
\label{app:task-memorisability}

 As a qualitative analysis, we categorise each of our chosen tasks into one of the Cognitive Skills categories from \citet{DBLP:journals/corr/abs-2301-06627}, since these categories may shed light on what kinds of linguistic and/or reasoning abilities are needed to understand a task. Additionally, we examine the degree of memorisability of each task. We define a task as \textit{memorisable} if a language model can conceivably achieve above-random performance on it by simply repeating factual information from its memory. Importantly, this would shortcut any reasoning path intended by the task, and performance would improve trivially as model size increases. Thus, we argue that performance gains on such tasks are unlikely to indicate emergence.\footnote{It is possible that, despite a task having high memorisability, a language model nevertheless goes through the intended reasoning process to arrive at the answer. In this case, a memorisable task could be considered emergent. But in this case, it would not be enough to merely show that performance improves with scale; one would also have to demonstrate that the language model is indeed reasoning. We forgo such an analysis here, and merely note that scale-related performance gains on highly-memorisable tasks are less likely to indicate emergence than non-memorisable tasks.}

In this section, we show memorisable and non-memorisable examples from each of our chosen tasks, to justify our evaluation of task memorisability from Section \ref{sec:results-emerge} (Emergence in GPT in the Absence of In-Context Learning), Table \ref{tab:default-results-ntz}. For tasks which contain no memorisable examples, or alternatively, no non-memorisable examples, the corresponding cell is left blank. A short explanation for the categorisation is provided below each example, in bold.

\begin{small}
\begin{longtblr}[
caption={Selected examples from each of our chosen tasks to justify our classification of memorisable vs. non-memorisable tasks. Note that some tasks contain both memorisable and non-memorisable examples, which occur in varying ratios as shown in Table \ref{tab:default-results-ntz}. Additionally, for our categorisation, we assume that leakage of task data is not a factor, i.e., an example is memorisable if and only if it can be solved through memory recall of information. We assume that previous memorisation of the actual question-answer pair has not occurred.},
label={app:memorisability-examples}
]{
colspec={ @{} X[halign=l] X[1.5,halign=l] X[2.1,halign=l] @{} },
row{1}={halign=l},
rowhead = 1
}
\toprule
\textbf{Task} & \textbf{Example Memorisable} & \textbf{Example Non-Memorisable} \\ \midrule
Causal judgement & n/a & The CEO of a company is sitting in his office when his Vice President of R\&D comes in and says, ``We are thinking of starting a new programme. It will help us increase profits, but it will also harm the environment.'' The CEO responds that he doesn't care about harming the environment and just wants to make as much profit as possible. The programme is carried out, profits are made and the environment is harmed. Did the CEO intentionally harm the environment?
\newline\textbf{\textit{Reason: Human-aligned moral reasoning necessary.}}
\\
\midrule
English Proverbs   & n/a & Vanessa spent lots of years helping out on weekends at the center for homeless aid. Recently, when she lost her job, the center was ready to offer a new job right away. Which of the following proverbs best apply to this situation?
\newline\textbf{\textit{Reason: Must connect a known proverb to a novel situation.}} \\
\midrule
Implicatures     & n/a & Speaker 1: ``Do you want to quit?'' \newline Speaker 2: ``I've never been the type of person who throws in the towel when things get tough.''
\newline\textbf{\textit{Reason: Pragmatics reasoning necessary.}}  \\
\midrule
Nonsense words grammar  & Which word in the following sentence is a verb? The grilshaws bolheavened whincely.
\newline\textbf{\textit{Reason: Linguistically-typical suffixes (i.e. -ed for a verb).}}
& Which word in the following sentence is a verb? I'd gralsillit onto the secure felisheret.
\newline\textbf{\textit{Reason: Linguistically-atypical suffixes (i.e. -it for a verb).}} \\
\midrule
Rhyming    & What rhymes with 'cruise'?
\newline\textbf{\textit{Reason: Model cannot rely on spelling or audio; rhyme dictionary knowledge necessary.}} & n/a     \\
\midrule
Tracking shuffled objects & n/a & Alice, Bob, and Claire are playing a game. At the start of the game, they are each holding a ball: Alice has a orange ball, Bob has a white ball, and Claire has a blue ball...At the end of the game, Alice has the?
\newline\textbf{\textit{Reason: Novel scenarios; state tracking abilities necessary.}} \\
\midrule
Commonsense QA         & Google Maps and other highway and street GPS services have replaced what?
\newline\textbf{\textit{Reason: Model can extract the answer from memorised articles about GPS services.}} & Sammy wanted to go to where the people were. Where might he go?
\newline\textbf{\textit{Reason: A novel, hypothetical scenario.}} \\
\midrule
GSM8K    & n/a & Natalia sold clips to 48 of her friends in April, and then she sold half as many clips in May. How many clips did Natalia sell altogether in April and May?
\newline\textbf{\textit{Reason: A novel question; math reasoning necessary.}} \\
\midrule
Analytic entailment  & \textit{The Great Gatsby} is a book written by F. Scott Fitzgerald. Therefore \textit{The Great Gatsby} comprises words.
\newline\textbf{\textit{Reason: Model can extract the fact that the book has words from an article describing the book.}}
& Tom is George’s grandfather. So, George is a descendant of Tom’s.
\newline\textbf{\textit{Reason: A novel, hypothetical scenario.}} \\
\midrule
Codenames   & Try to identify the 4 words best associated with the word DRIVE-IN from the following list...Give your answer in alphabetical order.
\newline\textbf{\textit{Reason: Model must determine word co-occurrence likelihood based on previously-encountered text.}} & n/a \\
\midrule
Common morpheme  & What is the common morpheme among these words: pyre, empyrean, antipyretic...
\newline\textbf{\textit{Reason: Model must determine word relations based on previously-encountered text.}} & n/a     \\
\midrule
Fact checker    & On June 2017, the following claim was made: The New Jersey Turnpike has zero shoulders. Was this claim true or false?
\newline\textbf{\textit{Reason: Model must recall information from previously-encountered text.}} & n/a \\
\midrule
Figure of speech detection & n/a & They fought like cats and dogs.
\newline\textbf{\textit{Reason: Model must determine the proper figurative language type of a novel sentence.}} \\
\midrule
Hindu knowledge   & Which of the following Hindu deities do not belong to the group of three supreme divinities known as the Trimurti?
\newline\textbf{\textit{Reason: Model must recall factual information about Hinduism.}}
& n/a          \\
\midrule
Logical deduction   & n/a & On a shelf, there are three books: a black book, an orange book, and a blue book. The blue book is to the right of the orange book. The orange book is to the right of the black book.
\newline\textbf{\textit{Reason: Model must keep track of spatially-oriented objects in novel scenarios.}}\\
\midrule
Misconceptions  & Twinkies are edible for decades or longer.
\newline\textbf{\textit{Reason: Model must recall factual information about common topics.}} & n/a \\
\midrule
Modified arithmetic & n/a & In the following lines, the symbol -> represents a simple mathematical operation. 
102 + 435 -> 537
...
466 + 214 ->
\newline\textbf{\textit{Reason: A novel question; math reasoning necessary.}}\\
\midrule
Phrase relatedness   & home town \newline ``town center'', ``location'', ``native city''...
\newline\textbf{\textit{Reason: Model must determine word co-occurrence likelihood based on previously-encountered text.}} & n/a \\
\midrule
Physical intuition  & An object is moving in a vacuum at velocity V with no net external forces acting on it. Does the object have nonzero acceleration?
\newline\textbf{\textit{Reason: Model must recall factual information about physics.}} & n/a \\
\midrule
Social IQA & n/a & Riley layered down their arms with a blanket. What does Riley need to do before this?
\newline\textbf{\textit{Reason: Model must reason about novel social situations.}}\\
\midrule
Strange stories    & n/a & Jane and Sarah are best friends. They both entered the same painting competition. Now Jane wanted to win this competition very much indeed, but when the results were announced it was her best friend Sarah who won, not her. Jane was very sad she had not won, but she was happy for her friend, who got the prize. Jane said to Sarah, ``Well done, I'm so happy you won!'' Jane said to her mother, ``I'm sad I didn't win that competition!'' Why does Jane say she is happy and sad at the same time?
\newline\textbf{\textit{Reason: Model must reason about novel social situations.}}        \\
\midrule
Strategy QA  & Was Pollock trained by Leonardo da Vinci?
\newline\textbf{\textit{Reason: Model can solve this by recalling previously-encountered text (such as a biography).}} & Could an escapee swim nonstop from Alcatraz island to Siberia?
\newline\textbf{\textit{Reason: Model must combine known concepts to a novel, hypothetical scenario.}}   \\
\bottomrule
\end{longtblr}
\end{small}

\newpage

\section{Complete results}
\label{app:completeresults}
In this section, we present our complete results. These encompass the performance plots for each of our 22 tasks, arranged in the following order by model type: GPT, T5, and  Other Models (Falcon and LLaMA). For each model, the results are ordered as follows: 
\begin{enumerate}
    \item Exact match accuracy in the closed prompt setting
    \item Exact match accuracy in the closed adversarial prompt setting
    \item Exact match accuracy in the open prompt setting
    \item BERTScore accuracy in the closed prompt setting
    \item BERTScore accuracy in the open prompt setting
    \item Edit distance in the closed prompt setting
    \item Edit distance in the open prompt setting
\end{enumerate}

Note that some metrics aren't compatible with all tasks (e.g., BERTScore accuracy with GSM8K, see Section \ref{sec:evalmetrics}), and that the \textit{codenames} task is incompatible with the open prompt setting, since the task requires choices to be provided in the input (see Section \ref{sec:evalmetrics} and Table \ref{app:memorisability-examples}). For this reason, some figures will contain fewer than 22 plots.

\begin{table}[h]
\small
\centering
\begin{tabular}{clll}
\toprule
\textbf{Model family}    & \textbf{Metric}                       & \textbf{Prompt format}                 & \textbf{Result}                                              \\ \midrule
\multirow{7}{*}{GPT}     & \multirow{3}{*}{Exact match accuracy} & closed                                 & Figure \ref{plot-gpt-exact_match_accuracy-closed.jpeg}       \\
                         &                                       & closed adversarial                     & Figure \ref{plot-gpt-exact_match_accuracy-closed-adv.jpeg}   \\
                         &                                       & open                                   & Figure \ref{plot-gpt-exact_match_accuracy-open.jpeg}         \\ \cmidrule{2-4}
                         & \multirow{2}{*}{BERTScore accuracy}   & closed                                 & Figure \ref{plot-gpt-bert_score_accuracy-closed.jpeg}        \\
                         &                                       & open                                   & Figure \ref{plot-gpt-bert_score_accuracy-open.jpeg}          \\ \cmidrule{2-4}
                         & \multirow{2}{*}{Edit distance}        & closed                                 & Figure \ref{plot-gpt-edit_distance-closed.jpeg}              \\
                         &                                       & open                                   & Figure \ref{plot-gpt-edit_distance-open.jpeg}                \\ \midrule
\multirow{7}{*}{T5}      & \multirow{3}{*}{Exact match accuracy} & closed                                 & Figure \ref{plot-t5-exact_match_accuracy-closed.jpeg}        \\
                         &                                       & closed adversarial                     & Figure \ref{plot-t5-exact_match_accuracy-closed-adv.jpeg}    \\
                         &                                       & open                                   & Figure \ref{plot-t5-exact_match_accuracy-open.jpeg}          \\ \cmidrule{2-4}
                         & \multirow{2}{*}{BERTScore accuracy}   & closed                                 & Figure \ref{plot-t5-bert_score_accuracy-closed.jpeg}         \\
                         &                                       & open                                   & Figure \ref{plot-t5-bert_score_accuracy-open.jpeg}           \\ \cmidrule{2-4}
                         & \multirow{2}{*}{Edit distance}        & closed                                 & Figure \ref{plot-t5-edit_distance-closed.jpeg}               \\
                         &                                       & open                                   & Figure \ref{plot-t5-edit_distance-open.jpeg}                 \\ \midrule
\multirow{7}{*}{Falcon}  & \multirow{3}{*}{Exact match accuracy} & closed                                 & Figure \ref{plot-falcon-exact_match_accuracy-closed.jpeg}    \\
                         &                                       & closed adversarial                     & Figure \ref{plot-falcon-exact_match_accuracy-closed-adv.jpeg}\\
                         &                                       & open                                   & Figure \ref{plot-falcon-exact_match_accuracy-open.jpeg}      \\ \cmidrule{2-4}
                         & \multirow{2}{*}{BERTScore accuracy}   & closed                                 & Figure \ref{plot-falcon-bert_score_accuracy-closed.jpeg}     \\
                         &                                       & open                                   & Figure \ref{plot-falcon-bert_score_accuracy-open.jpeg}       \\ \cmidrule{2-4}
                         & \multirow{2}{*}{Edit distance}        & closed                                 & Figure \ref{plot-falcon-edit_distance-closed.jpeg}           \\
                         &                                       & open                                   & Figure \ref{plot-falcon-edit_distance-open.jpeg}             \\ \midrule
\multirow{7}{*}{LLaMA}   & \multirow{3}{*}{Exact match accuracy} & closed                                 & Figure \ref{plot-llama-exact_match_accuracy-closed.jpeg}     \\
                         &                                       & closed adversarial                     & Figure \ref{plot-llama-exact_match_accuracy-closed-adv.jpeg} \\
                         &                                       & open                                   & Figure \ref{plot-llama-exact_match_accuracy-open.jpeg}       \\ \cmidrule{2-4}
                         & \multirow{2}{*}{BERTScore accuracy}   & closed                                 & Figure \ref{plot-llama-bert_score_accuracy-closed.jpeg}      \\
                         &                                       & open                                   & Figure \ref{plot-llama-bert_score_accuracy-open.jpeg}        \\ \cmidrule{2-4}
                         & \multirow{2}{*}{Edit distance}        & closed                                 & Figure \ref{plot-llama-edit_distance-closed.jpeg}            \\
                         &                                       & open                                   & Figure \ref{plot-llama-edit_distance-open.jpeg}              \\
\bottomrule
\end{tabular}
\caption{Performance plots (Result) for models in each model family (Model family) using different metrics (Metric) in the closed, closed adversarial, and open settings (Prompt format).}
\end{table}

\pagenumbering{gobble}

\begin{figure*}[htp]
\centering
\includegraphics[width=.92\textwidth]{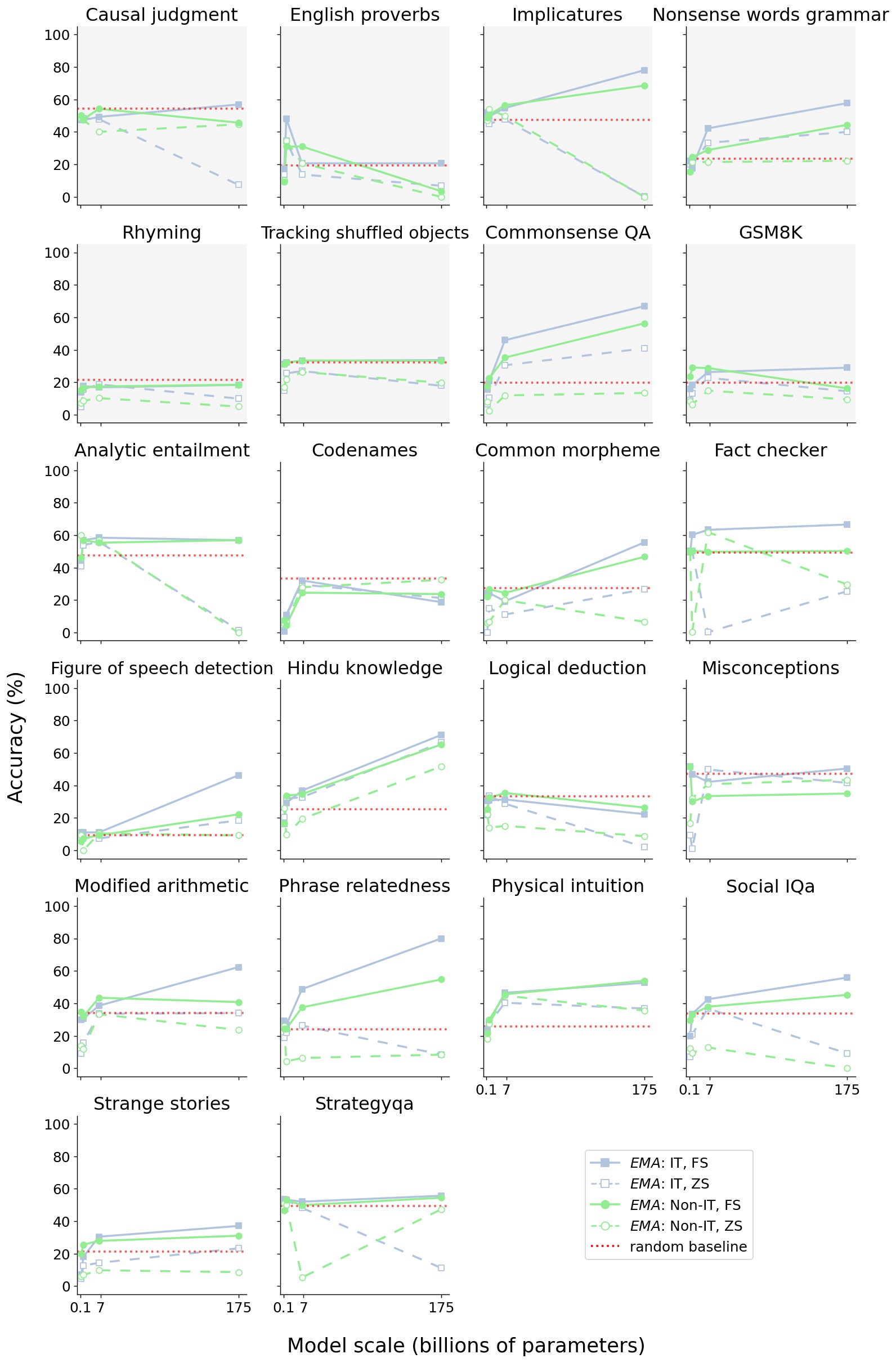}
\caption{Exact match accuracy (EMA) for instruction-tuned (IT) and non-instruction-tuned (Non-IT) GPT models using the closed prompt in the settings of zero-shot (ZS) and few-shot (FS).}
\label{plot-gpt-exact_match_accuracy-closed.jpeg}
\end{figure*}

\begin{figure*}[htp]
\centering
\includegraphics[width=.92\textwidth]{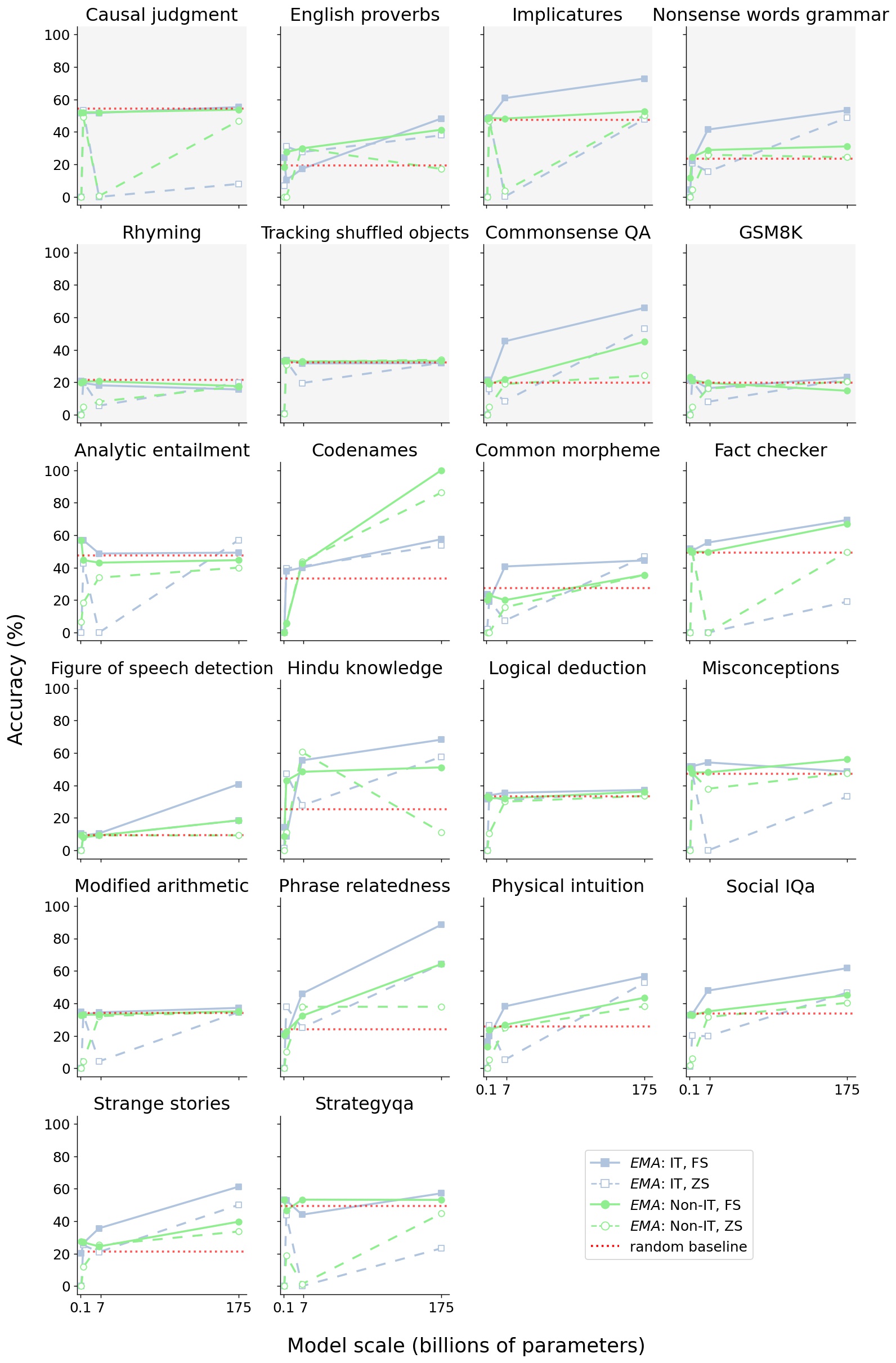}
\caption{Exact match accuracy (EMA) for instruction-tuned (IT) and non-instruction-tuned (Non-IT) GPT models using the closed adversarial prompt in the settings of zero-shot (ZS) and few-shot (FS).}
\label{plot-gpt-exact_match_accuracy-closed-adv.jpeg}
\end{figure*}

\begin{figure*}[htp]
\centering
\includegraphics[width=.92\textwidth]{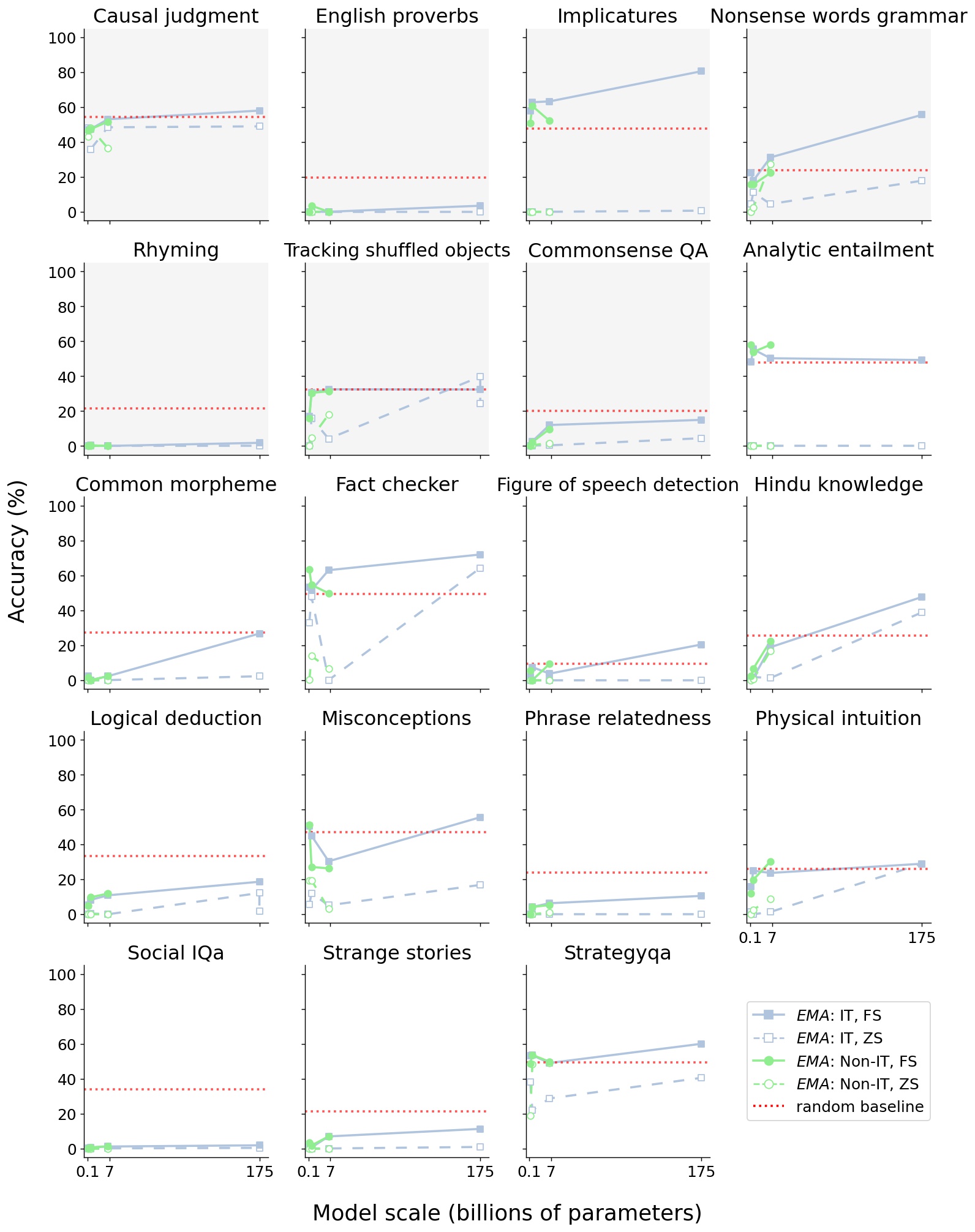}
\caption{Exact match accuracy (EMA) for instruction-tuned (IT) and non-instruction-tuned (Non-IT) GPT models using the open prompt in the settings of zero-shot (ZS) and few-shot (FS).}
\label{plot-gpt-exact_match_accuracy-open.jpeg}
\end{figure*}

\begin{figure*}[htp]
\centering
\includegraphics[width=.92\textwidth]{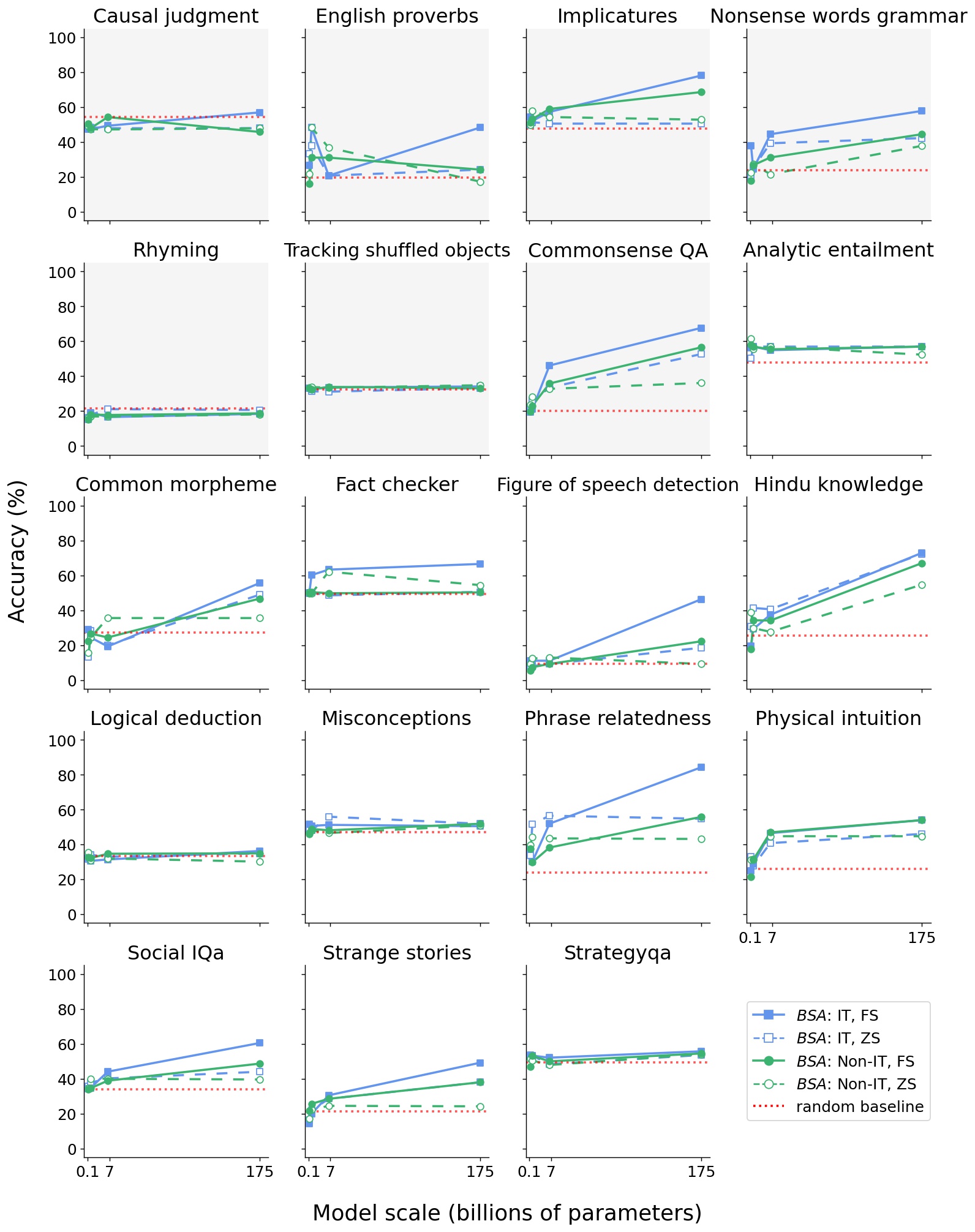}
\caption{BERTScore accuracy (BSA) for instruction-tuned (IT) and non-instruction-tuned (Non-IT) GPT models using the closed prompt in the settings of zero-shot (ZS) and few-shot (FS).}
\label{plot-gpt-bert_score_accuracy-closed.jpeg}
\end{figure*}

\begin{figure*}[htp]
\centering
\includegraphics[width=.92\textwidth]{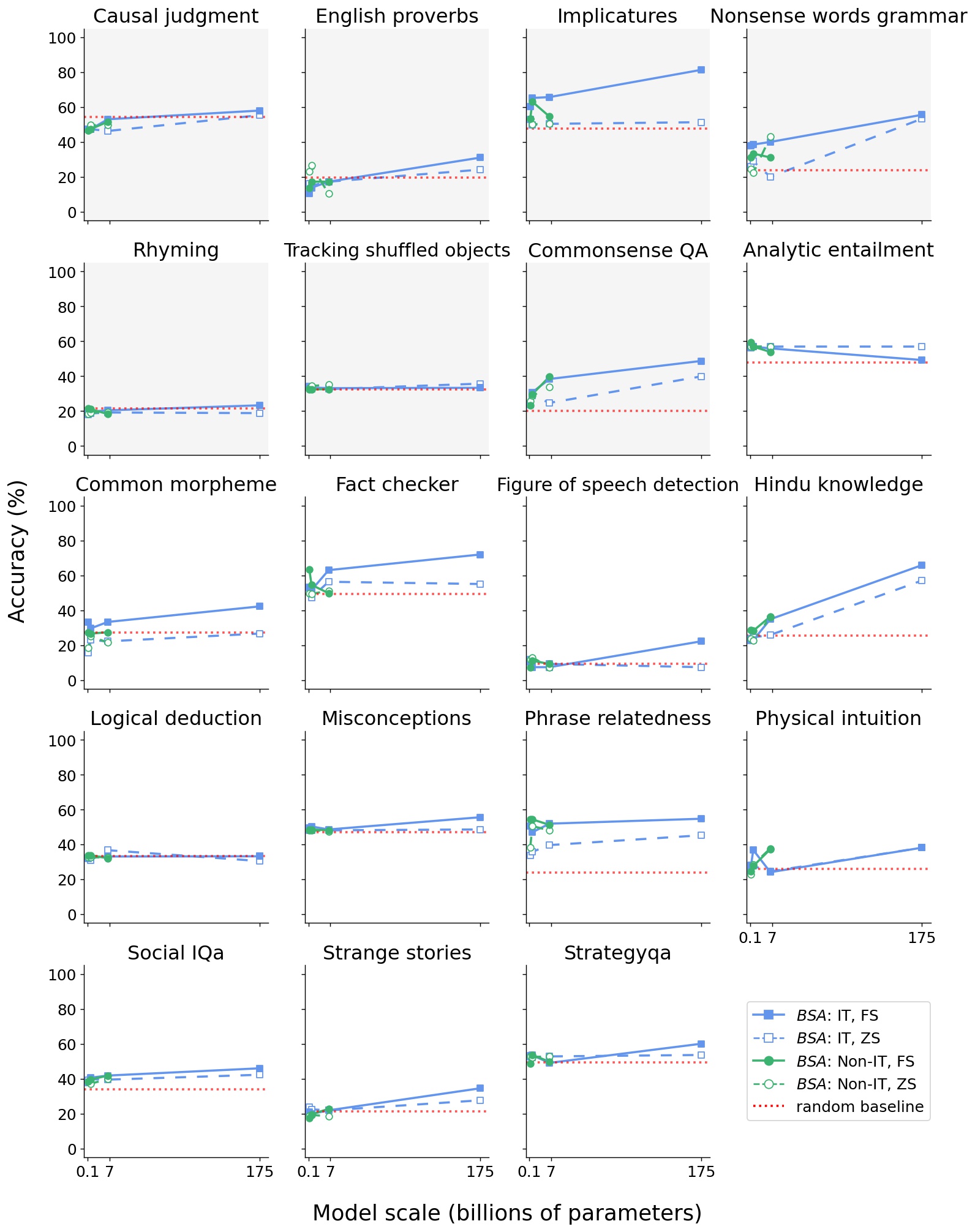}
\caption{BERTScore accuracy (BSA) for instruction-tuned (IT) and non-instruction-tuned (Non-IT) GPT models using the open prompt in the settings of zero-shot (ZS) and few-shot (FS).}
\label{plot-gpt-bert_score_accuracy-open.jpeg}
\end{figure*}

\begin{figure*}[htp]
\centering
\includegraphics[width=.92\textwidth]{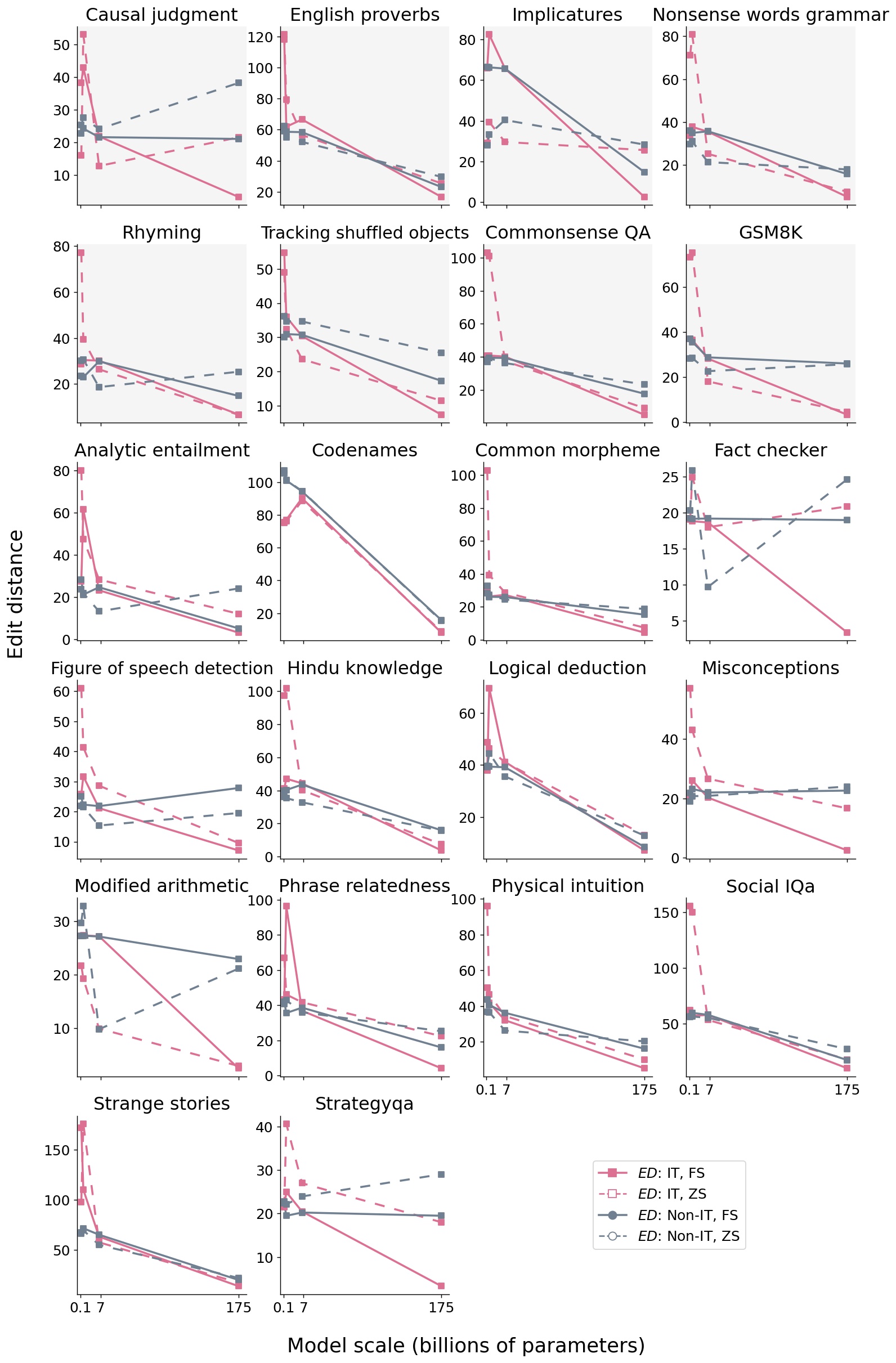}
\caption{Edit distance (ED) for instruction-tuned (IT) and non-instruction-tuned (Non-IT) GPT models using the closed prompt in the settings of zero-shot (ZS) and few-shot (FS).}
\label{plot-gpt-edit_distance-closed.jpeg}
\end{figure*}

\begin{figure*}[htp]
\centering
\includegraphics[width=.92\textwidth]{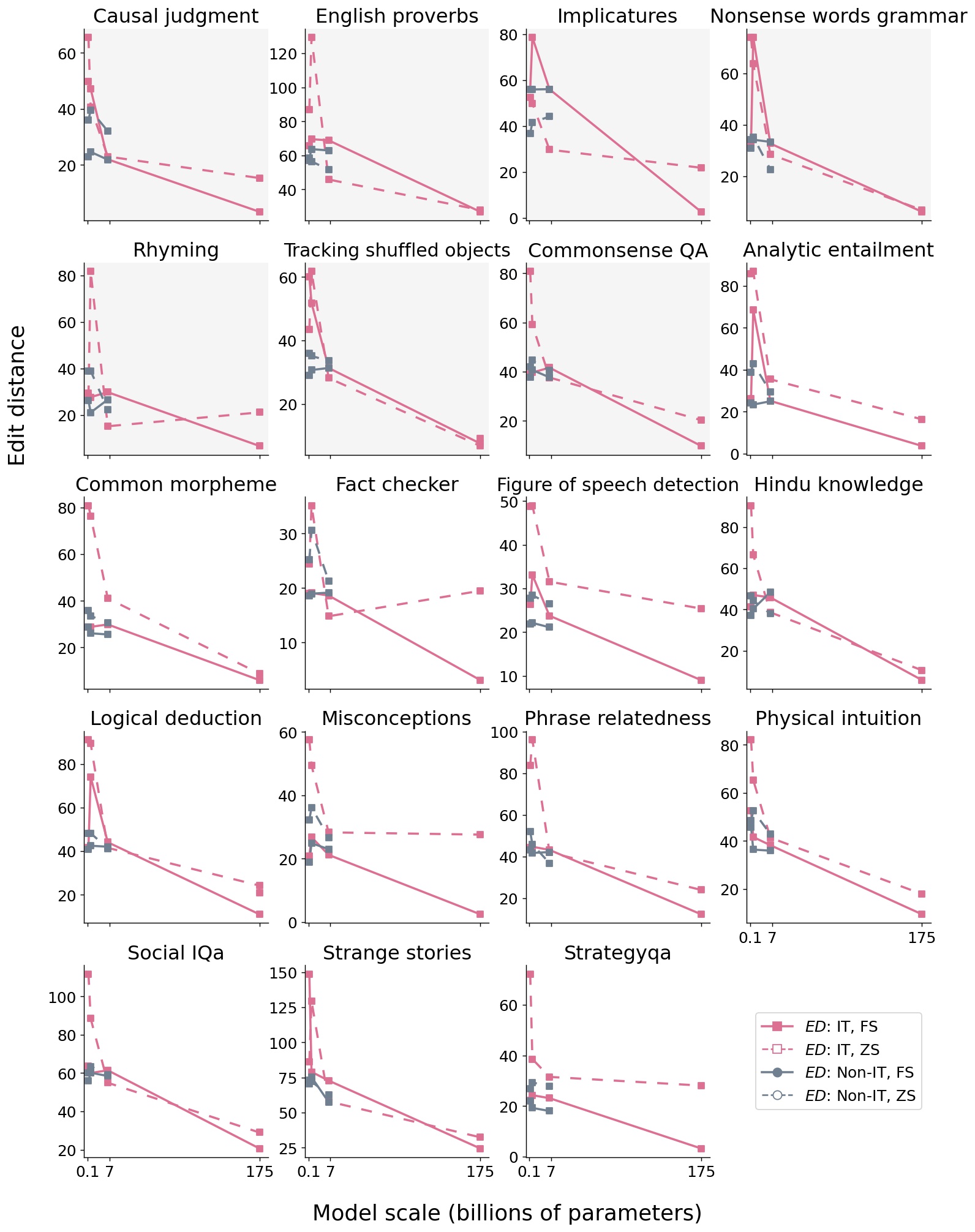}
\caption{Edit distance (ED) for instruction-tuned (IT) and non-instruction-tuned (Non-IT) GPT models using the open prompt in the settings of zero-shot (ZS) and few-shot (FS).}
\label{plot-gpt-edit_distance-open.jpeg}
\end{figure*}

\newpage

\begin{figure*}[]
\centering
\includegraphics[width=.92\textwidth]{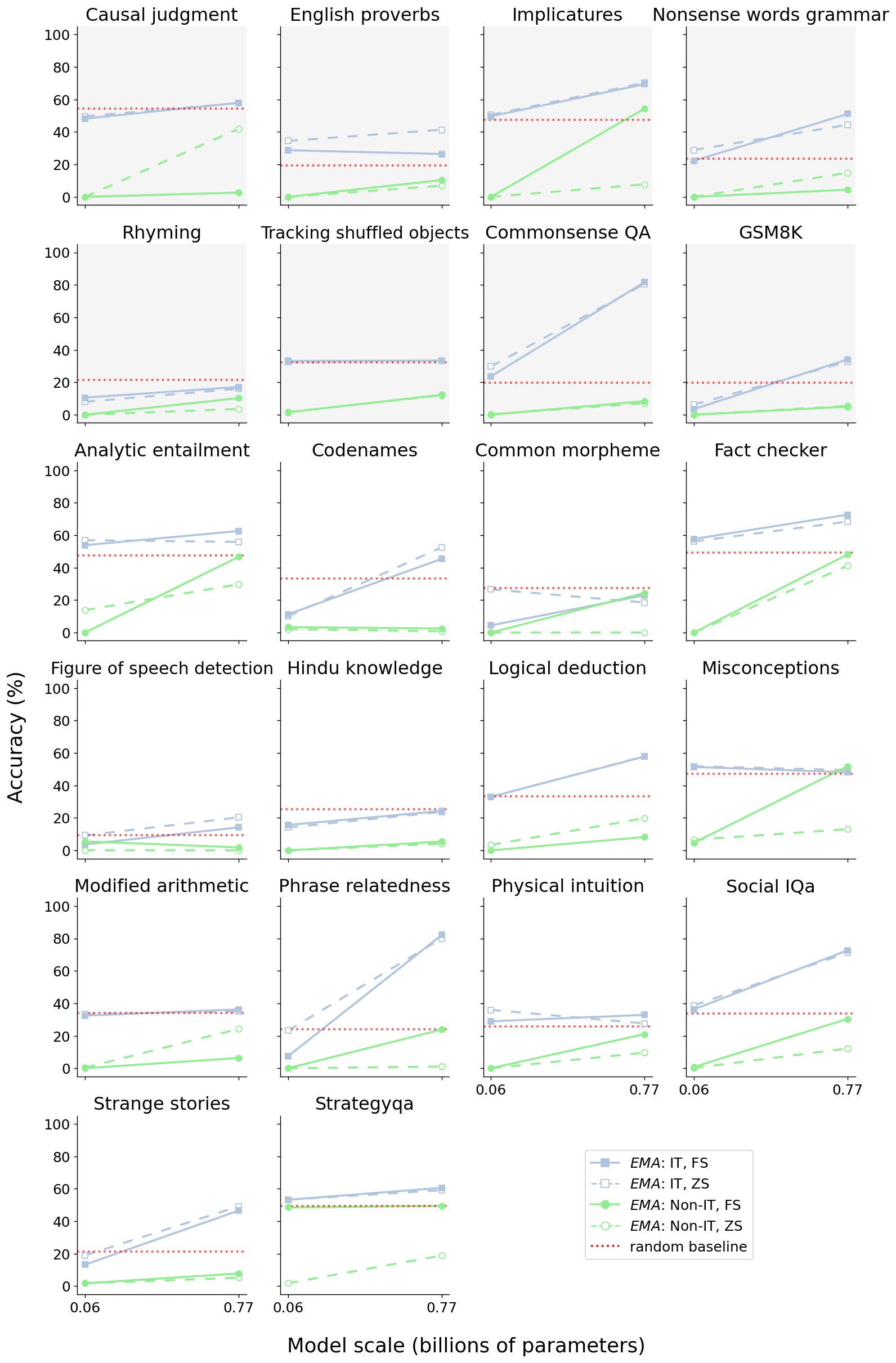}
\caption{Exact match accuracy (EMA) for instruction-tuned (IT) and non-instruction-tuned (Non-IT) T5 models using the closed prompt in the settings of zero-shot (ZS) and few-shot (FS).}
\label{plot-t5-exact_match_accuracy-closed.jpeg}
\end{figure*}

\begin{figure*}[]
\centering
\includegraphics[width=.92\textwidth]{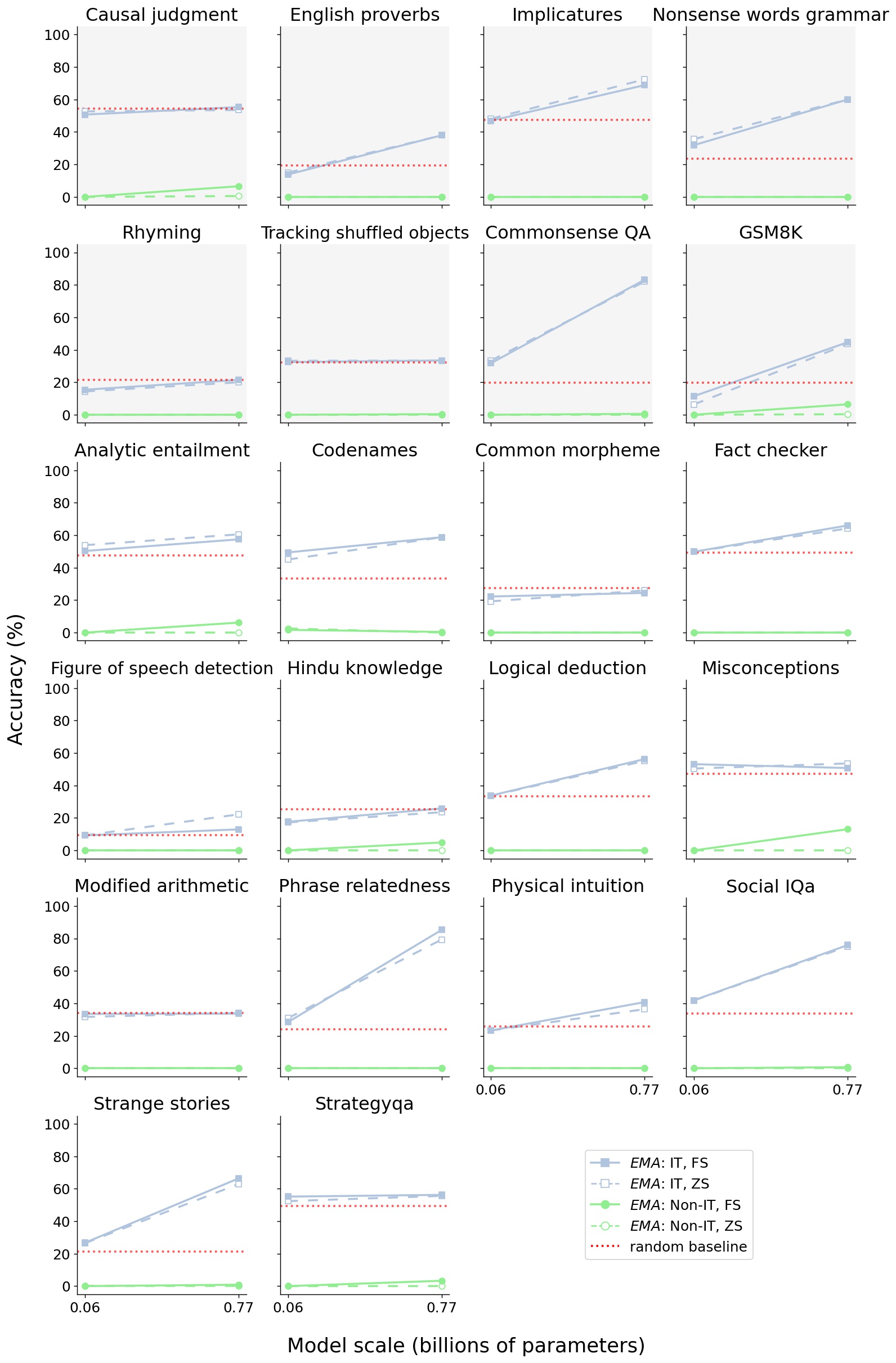}
\caption{Exact match accuracy (EMA) for instruction-tuned (IT) and non-instruction-tuned (Non-IT) T5 models using the closed adversarial prompt in the settings of zero-shot (ZS) and few-shot (FS).}
\label{plot-t5-exact_match_accuracy-closed-adv.jpeg}
\end{figure*}

\begin{figure*}[]
\centering
\includegraphics[width=.92\textwidth]{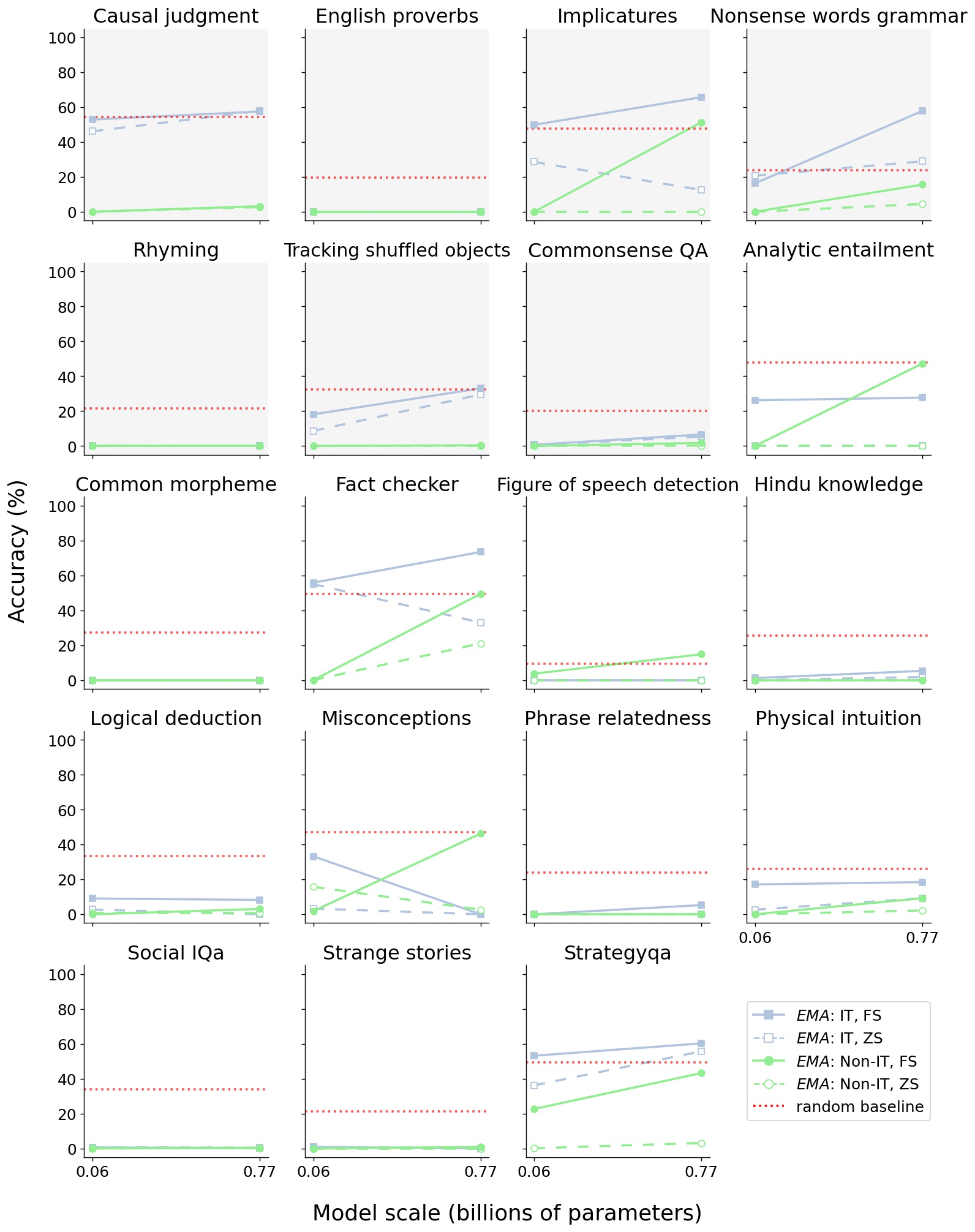}
\caption{Exact match accuracy (EMA) for instruction-tuned (IT) and non-instruction-tuned (Non-IT) T5 models using the open prompt in the settings of zero-shot (ZS) and few-shot (FS).}
\label{plot-t5-exact_match_accuracy-open.jpeg}
\end{figure*}

\begin{figure*}[]
\centering
\includegraphics[width=.92\textwidth]{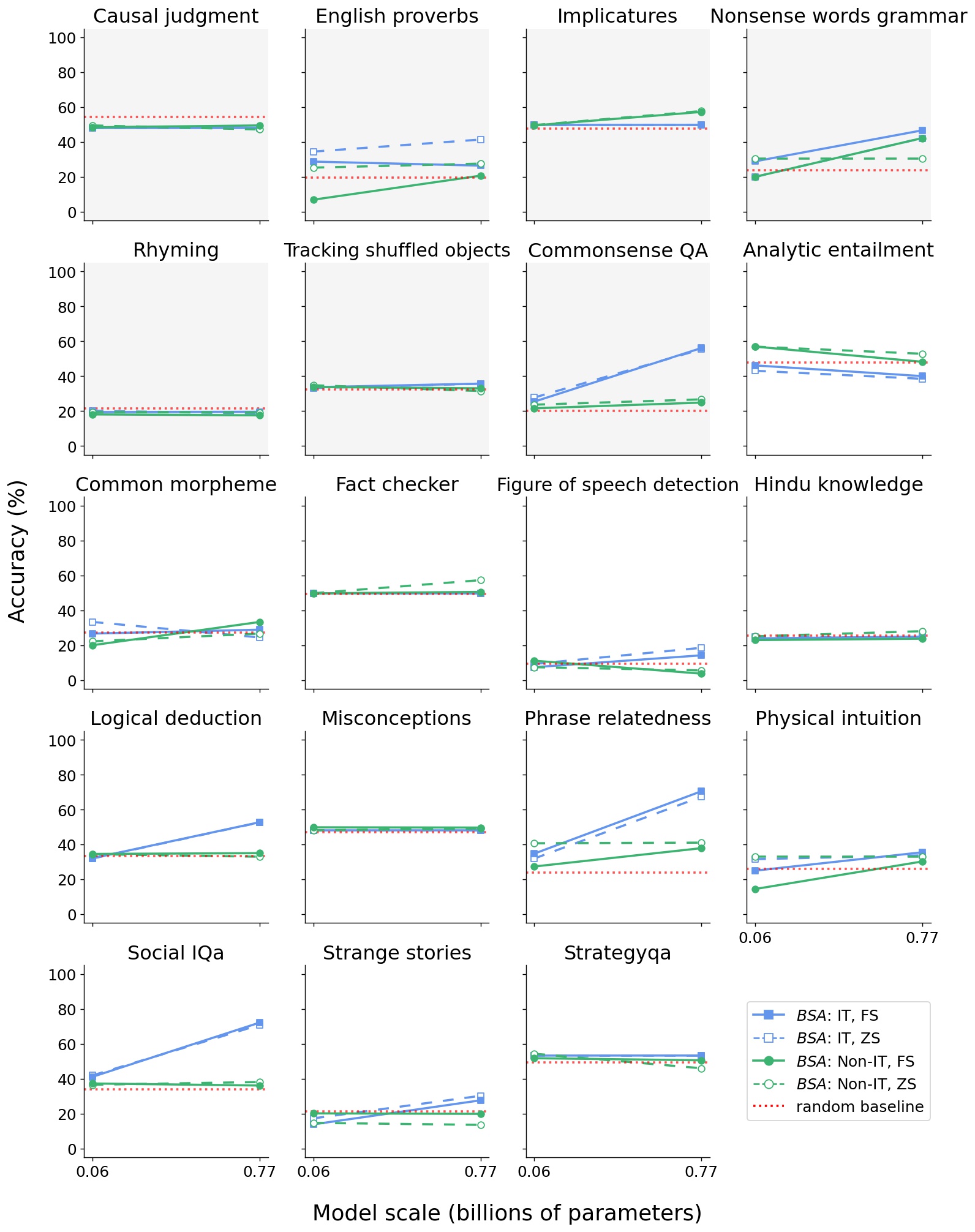}
\caption{BERTScore accuracy (BSA) for instruction-tuned (IT) and non-instruction-tuned (Non-IT) T5 models using the closed prompt in the settings of zero-shot (ZS) and few-shot (FS).}
\label{plot-t5-bert_score_accuracy-closed.jpeg}
\end{figure*}

\begin{figure*}[]
\centering
\includegraphics[width=.92\textwidth]{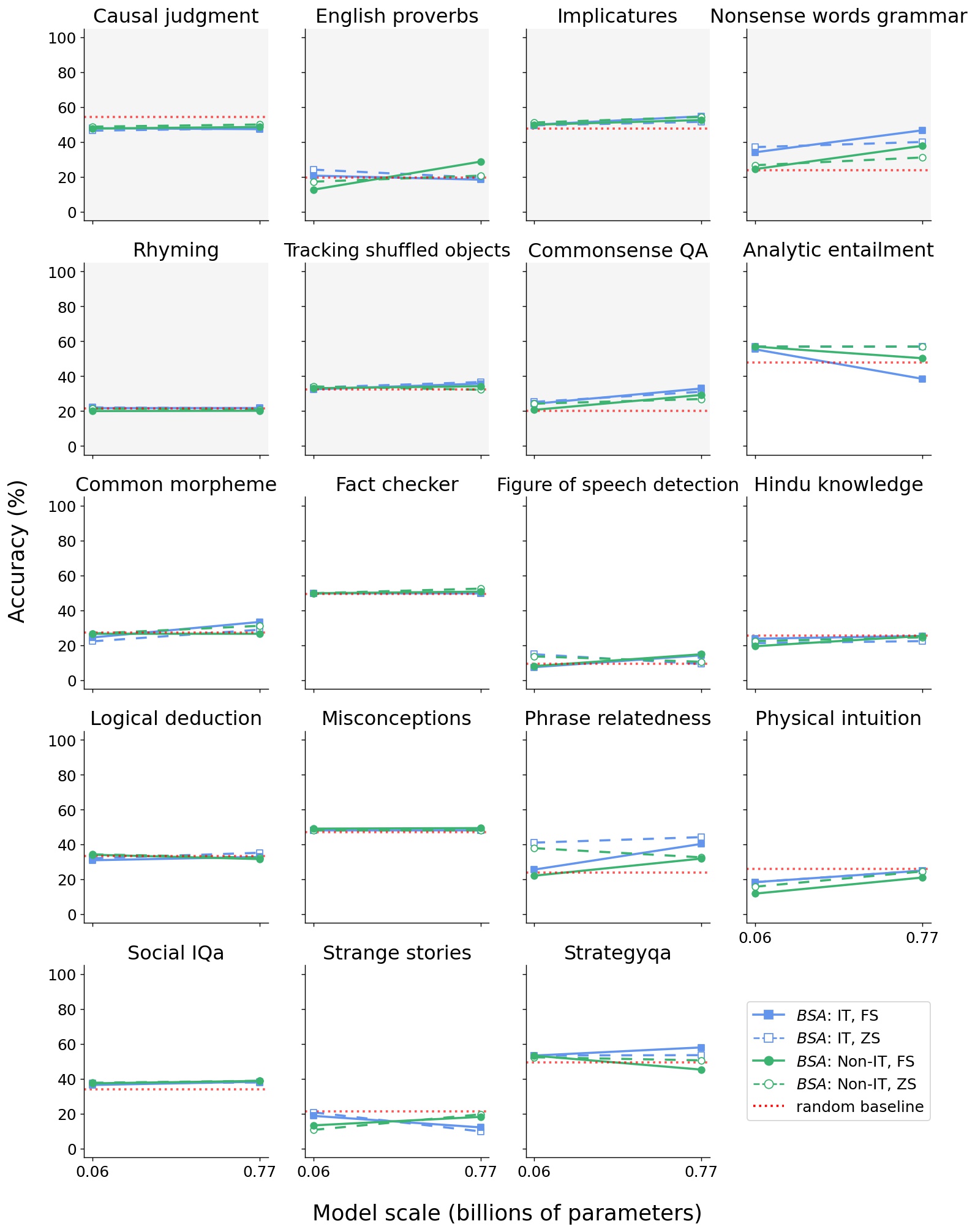}
\caption{BERTScore accuracy (BSA) for instruction-tuned (IT) and non-instruction-tuned (Non-IT) T5 models using the open prompt in the settings of zero-shot (ZS) and few-shot (FS).}
\label{plot-t5-bert_score_accuracy-open.jpeg}
\end{figure*}

\begin{figure*}[]
\centering
\includegraphics[width=.92\textwidth]{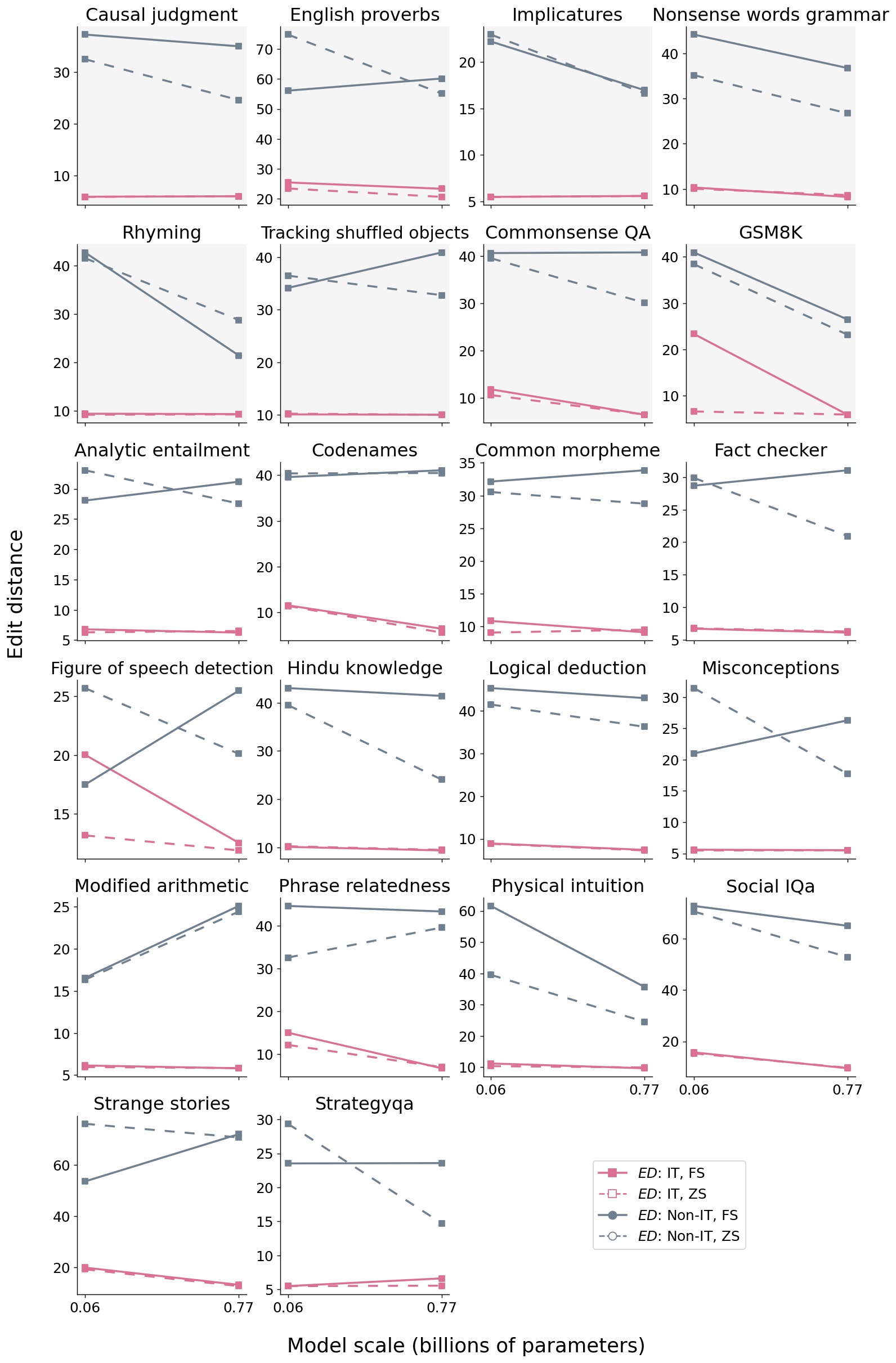}
\caption{Edit distance (ED) for instruction-tuned (IT) and non-instruction-tuned (Non-IT) T5 models using the closed prompt in the settings of zero-shot (ZS) and few-shot (FS).}
\label{plot-t5-edit_distance-closed.jpeg}
\end{figure*}

\begin{figure*}[]
\centering
\includegraphics[width=.92\textwidth]{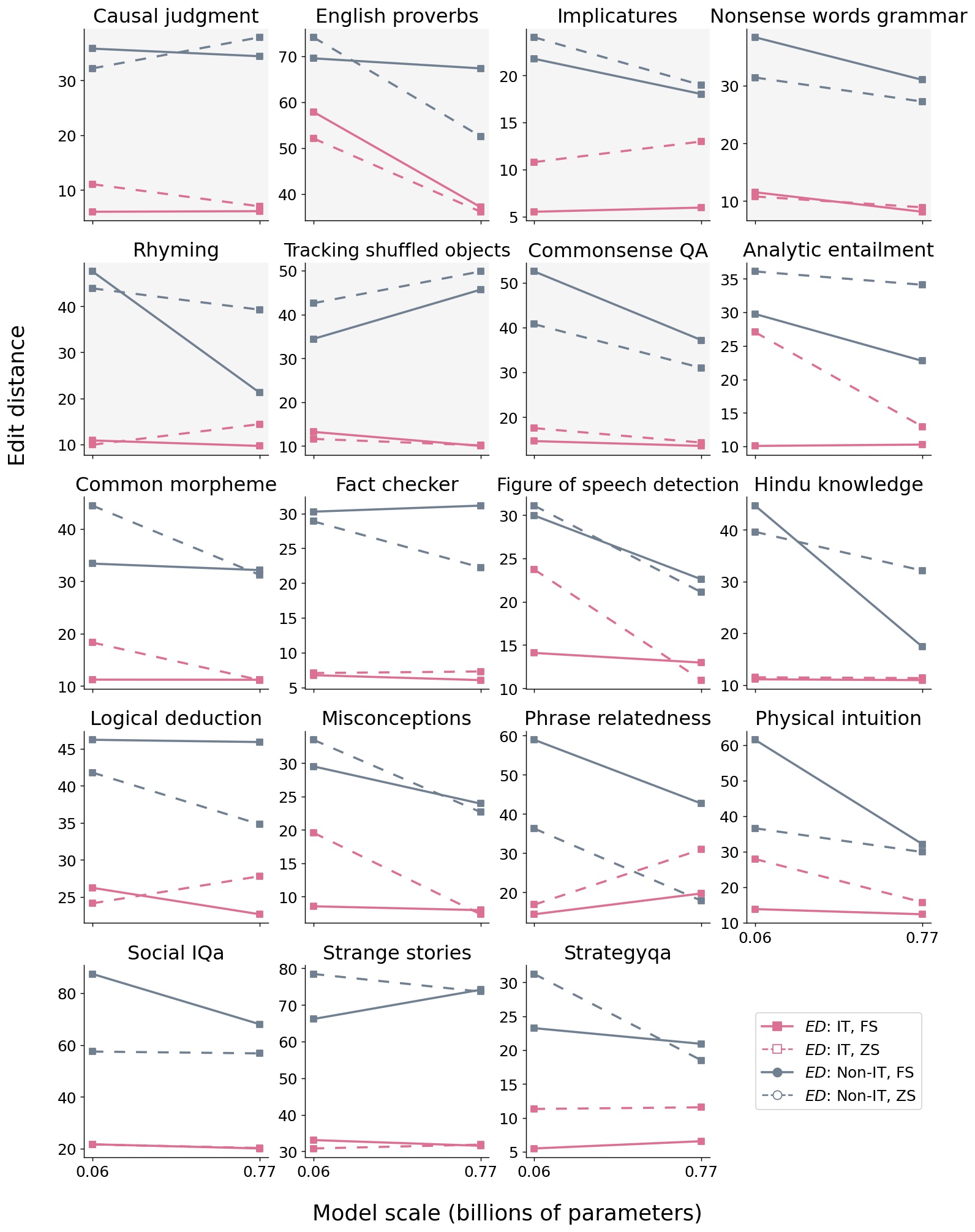}
\caption{Edit distance (ED) for instruction-tuned (IT) and non-instruction-tuned (Non-IT) T5 models using the open prompt in the settings of zero-shot (ZS) and few-shot (FS).}
\label{plot-t5-edit_distance-open.jpeg}
\end{figure*}

\newpage

\begin{figure*}[!htb]
\centering
\includegraphics[width=.92\textwidth]{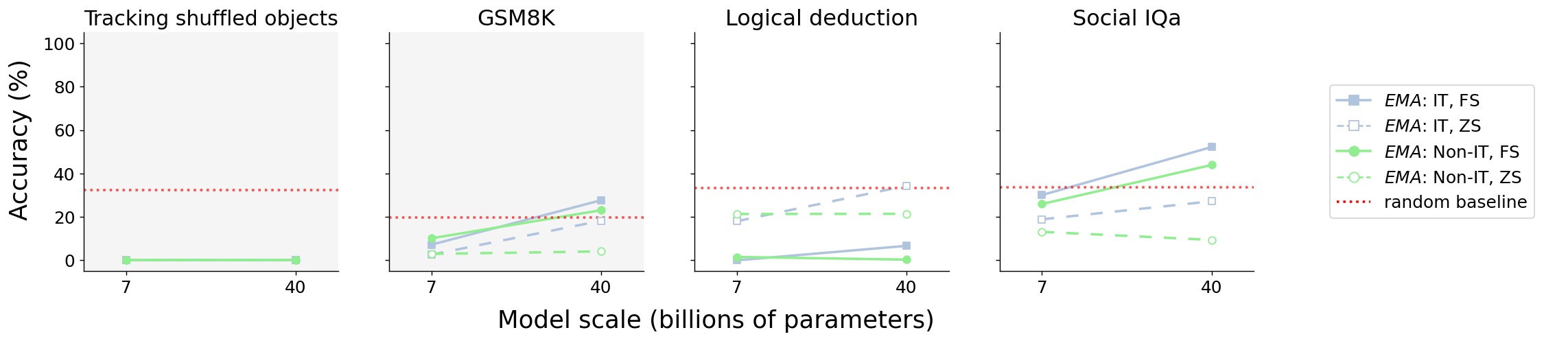}
\caption{Exact match accuracy (EMA) for instruction-tuned (IT) and non-instruction-tuned (Non-IT) Falcon models using the closed prompt in the settings of zero-shot (ZS) and few-shot (FS).}
\label{plot-falcon-exact_match_accuracy-closed.jpeg}
\end{figure*}

\begin{figure*}[!htb]
\centering
\includegraphics[width=.92\textwidth]{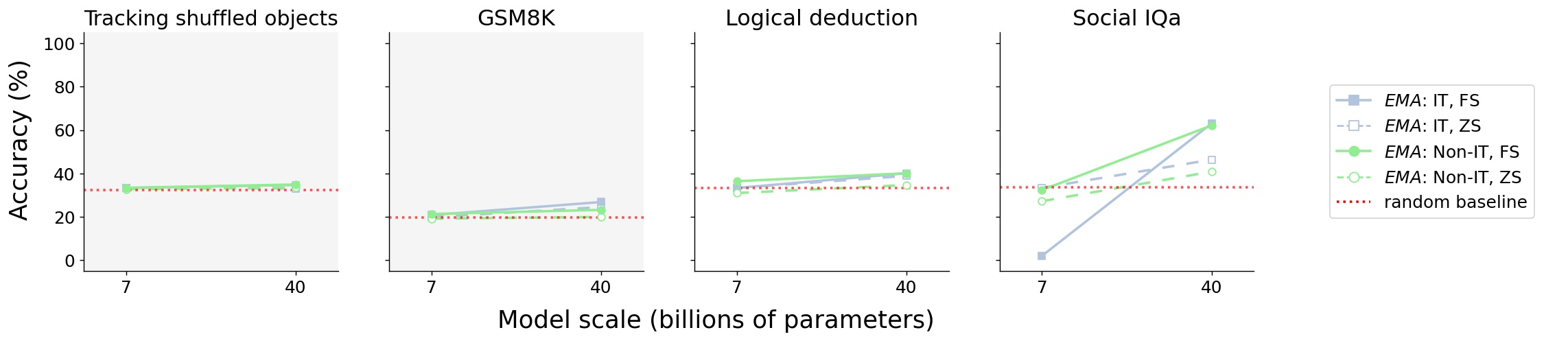}
\caption{Exact match accuracy (EMA) for instruction-tuned (IT) and non-instruction-tuned (Non-IT) Falcon models using the closed adversarial prompt in the settings of zero-shot (ZS) and few-shot (FS).}
\label{plot-falcon-exact_match_accuracy-closed-adv.jpeg}
\end{figure*}

\begin{figure*}[!htb]
\centering
\includegraphics[width=.92\textwidth]{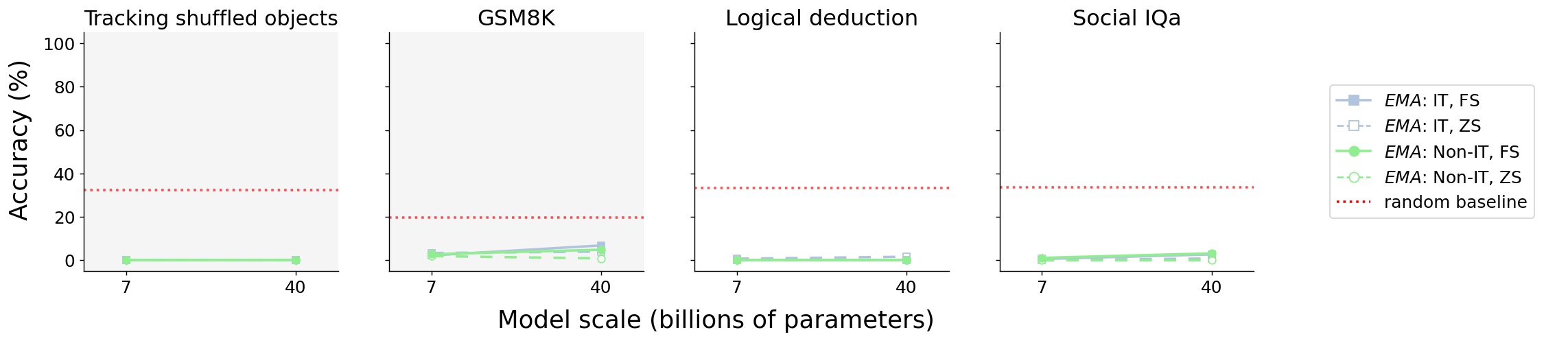}
\caption{Exact match accuracy (EMA) for instruction-tuned (IT) and non-instruction-tuned (Non-IT) Falcon models using the open prompt in the settings of zero-shot (ZS) and few-shot (FS).}
\label{plot-falcon-exact_match_accuracy-open.jpeg}
\end{figure*}

\begin{figure*}[!htb]
\centering
\includegraphics[width=.92\textwidth]{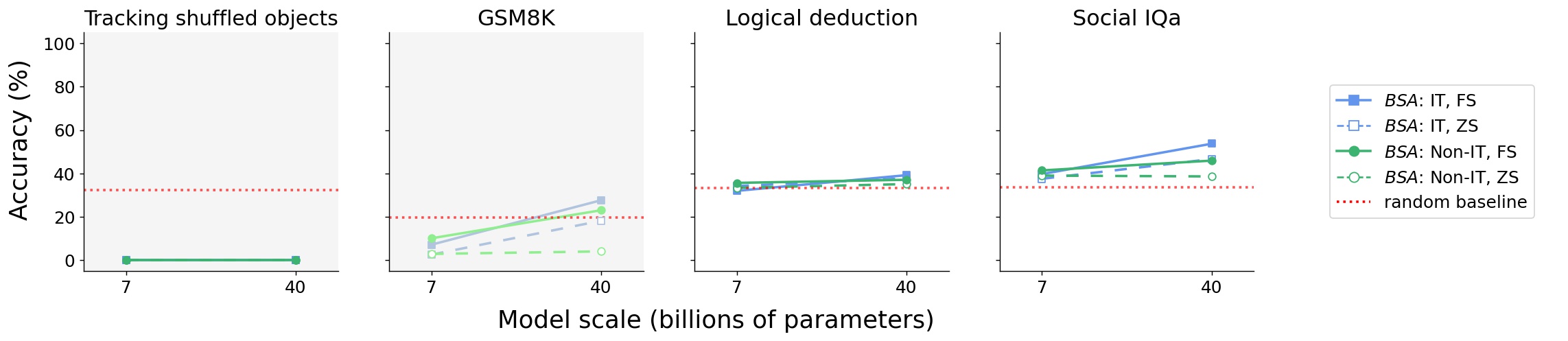}
\caption{BERTScore accuracy (BSA) for instruction-tuned (IT) and non-instruction-tuned (Non-IT) Falcon models using the closed prompt in the settings of zero-shot (ZS) and few-shot (FS).}
\label{plot-falcon-bert_score_accuracy-closed.jpeg}
\end{figure*}

\begin{figure*}[!htb]
\centering
\includegraphics[width=.92\textwidth]{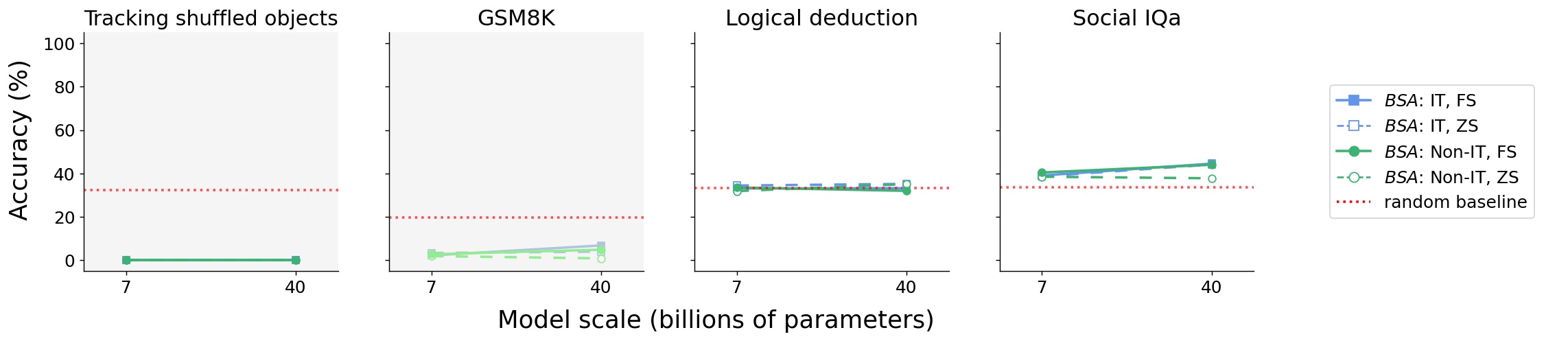}
\caption{BERTScore accuracy (BSA) for instruction-tuned (IT) and non-instruction-tuned (Non-IT) Falcon models using the open prompt in the settings of zero-shot (ZS) and few-shot (FS).}
\label{plot-falcon-bert_score_accuracy-open.jpeg}
\end{figure*}

\begin{figure*}[!htb]
\centering
\includegraphics[width=.92\textwidth]{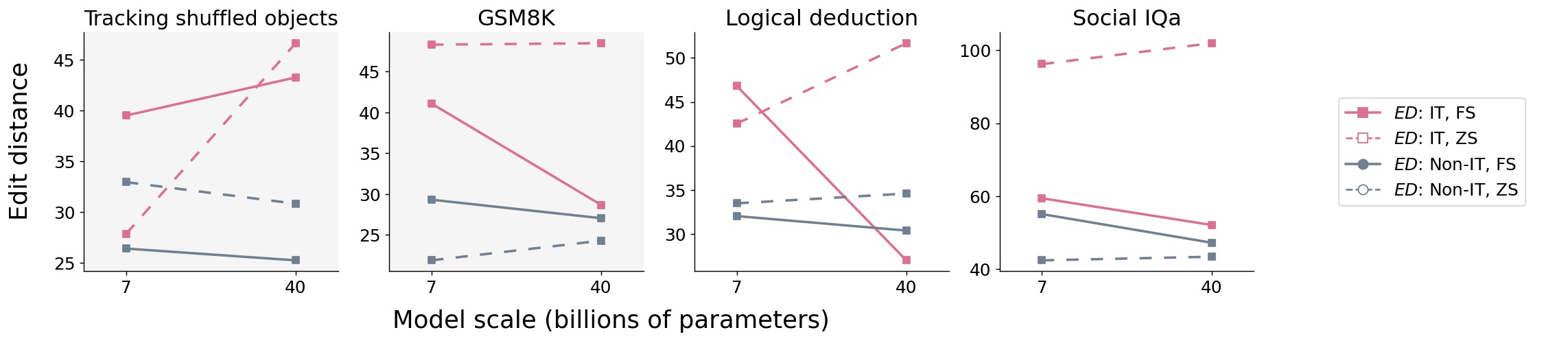}
\caption{Edit distance (ED) for instruction-tuned (IT) and non-instruction-tuned (Non-IT) Falcon models using the closed prompt in the settings of zero-shot (ZS) and few-shot (FS).}
\label{plot-falcon-edit_distance-closed.jpeg}
\end{figure*}

\begin{figure*}[!htb]
\centering
\includegraphics[width=.92\textwidth]{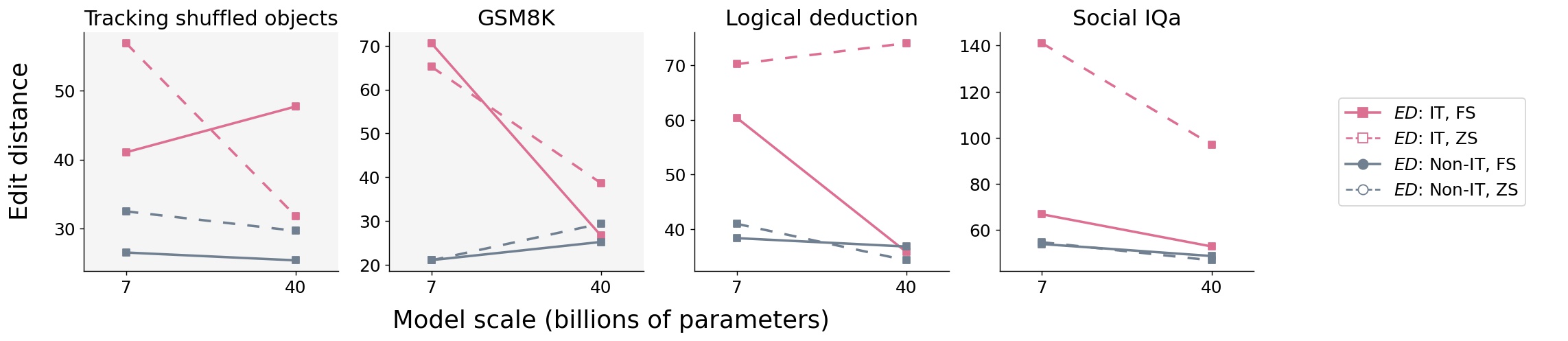}
\caption{Edit distance (ED) for instruction-tuned (IT) and non-instruction-tuned (Non-IT) Falcon models using the open prompt in the settings of zero-shot (ZS) and few-shot (FS).}
\label{plot-falcon-edit_distance-open.jpeg}
\end{figure*}

\begin{figure*}[!htb]
\centering
\includegraphics[width=.92\textwidth]{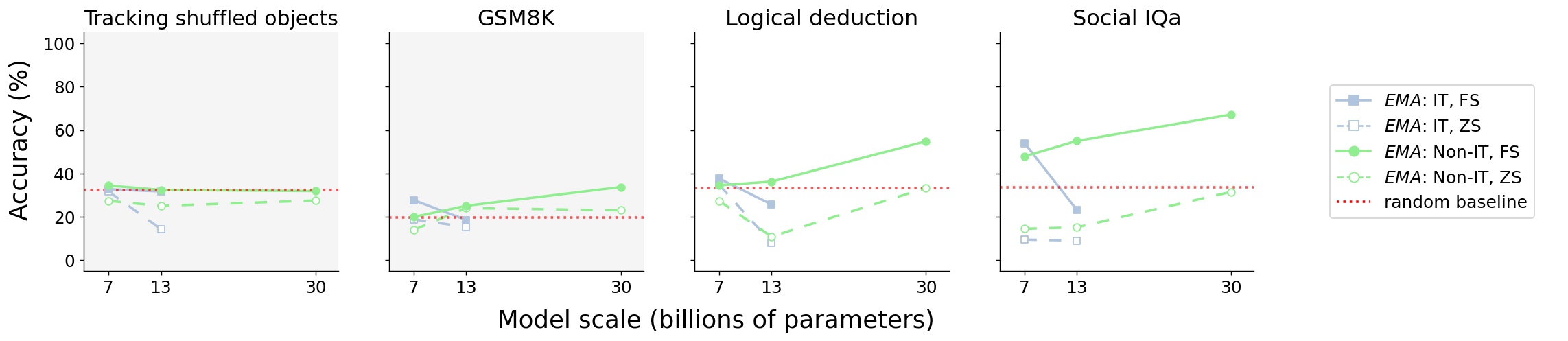}
\caption{Exact match accuracy (EMA) for instruction-tuned (IT) and non-instruction-tuned (Non-IT) LLaMA models using the closed prompt in the settings of zero-shot (ZS) and few-shot (FS).}
\label{plot-llama-exact_match_accuracy-closed.jpeg}
\end{figure*}

\begin{figure*}[!htb]
\centering
\includegraphics[width=.92\textwidth]{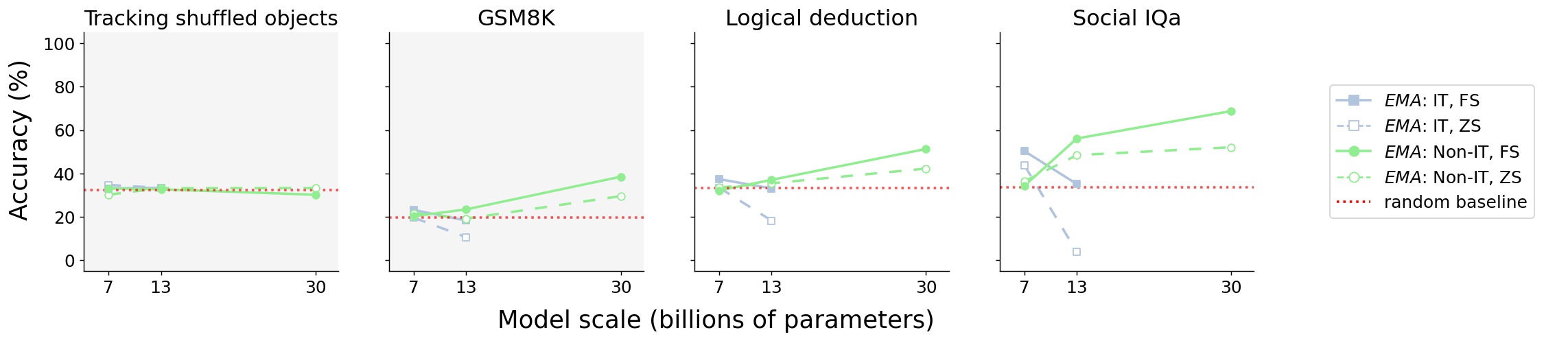}
\caption{Exact match accuracy (EMA) for instruction-tuned (IT) and non-instruction-tuned (Non-IT) LLaMA models using the closed adversarial prompt in the settings of zero-shot (ZS) and few-shot (FS).}
\label{plot-llama-exact_match_accuracy-closed-adv.jpeg}
\end{figure*}

\begin{figure*}[!htb]
\centering
\includegraphics[width=.92\textwidth]{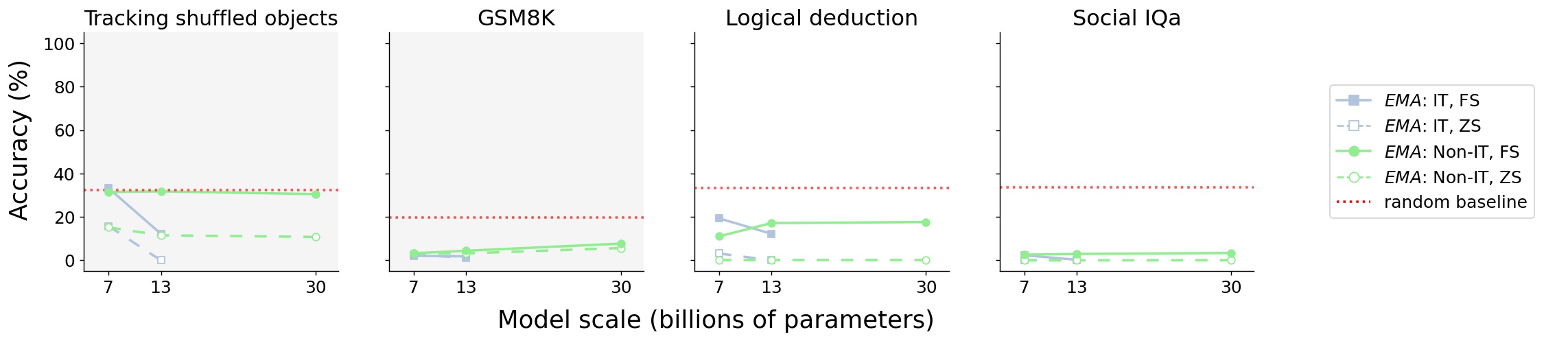}
\caption{Exact match accuracy (EMA) for instruction-tuned (IT) and non-instruction-tuned (Non-IT) LLaMA models using the open prompt in the settings of zero-shot (ZS) and few-shot (FS).}
\label{plot-llama-exact_match_accuracy-open.jpeg}
\end{figure*}

\begin{figure*}[!htb]
\centering
\includegraphics[width=.92\textwidth]{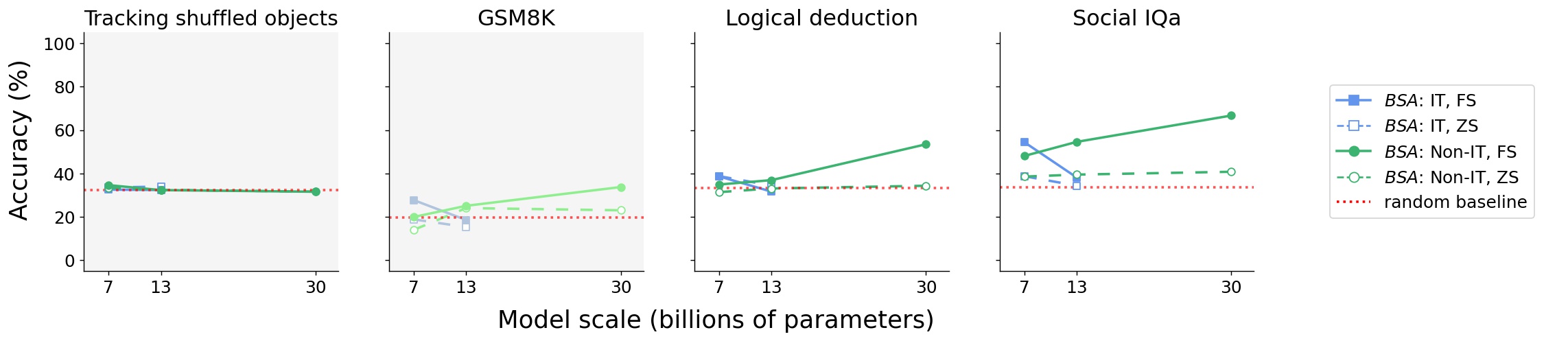}
\caption{BERTScore accuracy (BSA) for instruction-tuned (IT) and non-instruction-tuned (Non-IT) LLaMA models using the closed prompt in the settings of zero-shot (ZS) and few-shot (FS).}
\label{plot-llama-bert_score_accuracy-closed.jpeg}
\end{figure*}

\begin{figure*}[!htb]
\centering
\includegraphics[width=.92\textwidth]{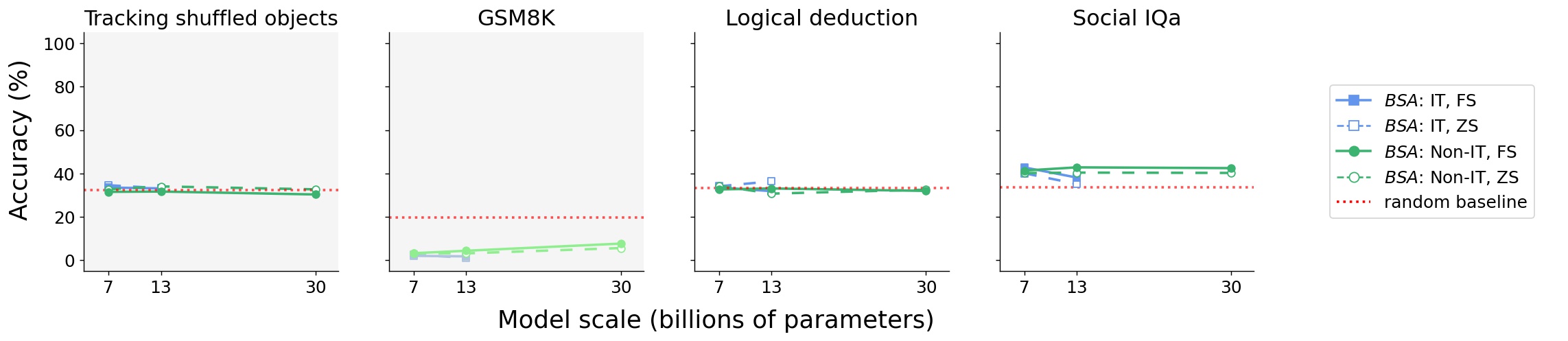}
\caption{BERTScore accuracy (BSA) for instruction-tuned (IT) and non-instruction-tuned (Non-IT) LLaMA models using the open prompt in the settings of zero-shot (ZS) and few-shot (FS).}
\label{plot-llama-bert_score_accuracy-open.jpeg}
\end{figure*}

\begin{figure*}[!htb]
\centering
\includegraphics[width=.92\textwidth]{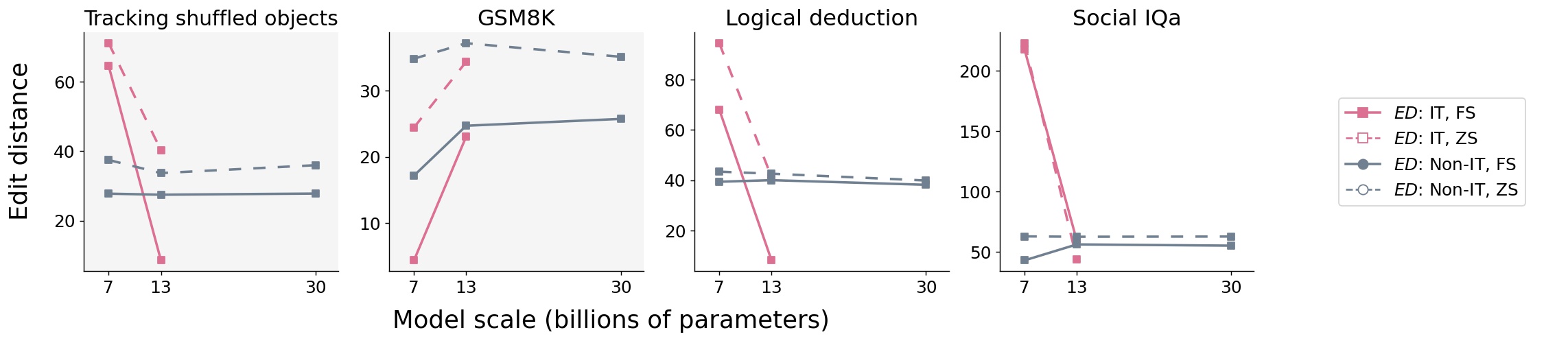}
\caption{Edit distance (ED) for instruction-tuned (IT) and non-instruction-tuned (Non-IT) LLaMA models using the closed prompt in the settings of zero-shot (ZS) and few-shot (FS).}
\label{plot-llama-edit_distance-closed.jpeg}
\end{figure*}

\begin{figure*}[!htb]
\centering
\includegraphics[width=.92\textwidth]{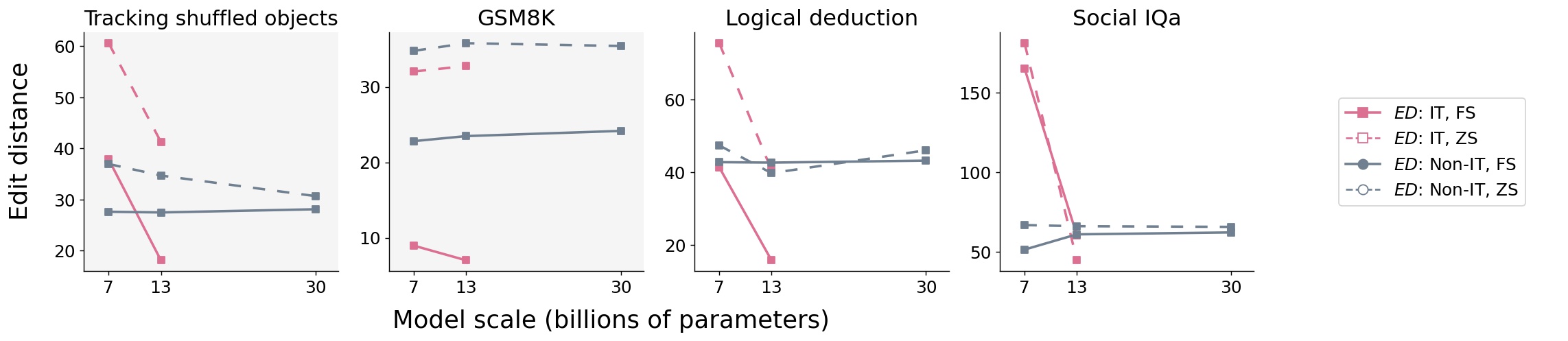}
\caption{Edit distance (ED) for instruction-tuned (IT) and non-instruction-tuned (Non-IT) LLaMA models using the open prompt in the settings of zero-shot (ZS) and few-shot (FS).}
\label{plot-llama-edit_distance-open.jpeg}
\end{figure*}

\end{document}